\pdfoutput=1

\documentclass[11pt]{article}

\usepackage[final]{acl}

\usepackage{times}
\usepackage{latexsym}
\usepackage{booktabs}
\usepackage{enumitem}
\usepackage{pifont}

\usepackage[T1]{fontenc}

\usepackage[utf8]{inputenc}

\usepackage{microtype}

\usepackage{inconsolata}

\usepackage{graphicx}

\usepackage{multicol}
\usepackage{multirow}
\usepackage{amsmath}
\usepackage{xcolor}
\usepackage{subfigure}
\newcommand{\review}[1]{\textcolor{black}{#1}}

\usepackage{multirow}
\usepackage[LFE,LAE,T1]{fontenc}
\usepackage[arabic, main=english]{babel}
\usepackage[textsize=tiny]{todonotes}

\newcommand{\allnotes}[1]{}
\renewcommand{\allnotes}[1]{#1} 

\title{Evaluating the Robustness and Accuracy of Text Watermarking Under Real-World Cross-Lingual Manipulations}

\author{
Mansour Al Ghanim \enskip Jiaqi Xue \enskip Rochana Prih Hastuti \enskip Mengxin Zheng \enskip Yan Solihin \enskip Qian Lou \\
University of Central Florida \\
\texttt{\{mansour.alghanim,jiaqi.xue,rochana,mengxin.zheng,yan.solihin,qian.lou\}@ucf.edu} \\
}

\begin{document}
\maketitle
\begingroup
\renewcommand\thefootnote{}
\footnotetext{Accepted to the Findings of EMNLP-2025}
\endgroup
\begin{abstract}
\review{We present a study to benchmark representative watermarking methods in cross-lingual settings. The current literature mainly focuses on the evaluation of watermarking methods for the English language. However, the literature for evaluating watermarking in cross-lingual settings is scarce. This results in overlooking important adversary scenarios in which a cross-lingual adversary could be in, leading to a gray area of practicality over cross-lingual watermarking. In this paper, we evaluate four watermarking methods in four different and vocabulary rich languages. Our experiments investigate the quality of text under different watermarking procedure and the detectability of watermarks with practical translation attack scenarios. Specifically, we investigate practical scenarios that an adversary with cross-lingual knowledge could take, and evaluate whether current watermarking methods are suitable for such scenarios. Finally, from our findings, we draw key insights about watermarking in cross-lingual settings\footnote{code and data: \url{https://github.com/SecureDL/xlingual_watermark_eval}}.}

\end{abstract}

\begin{figure*}
  \includegraphics[width=0.98\textwidth]{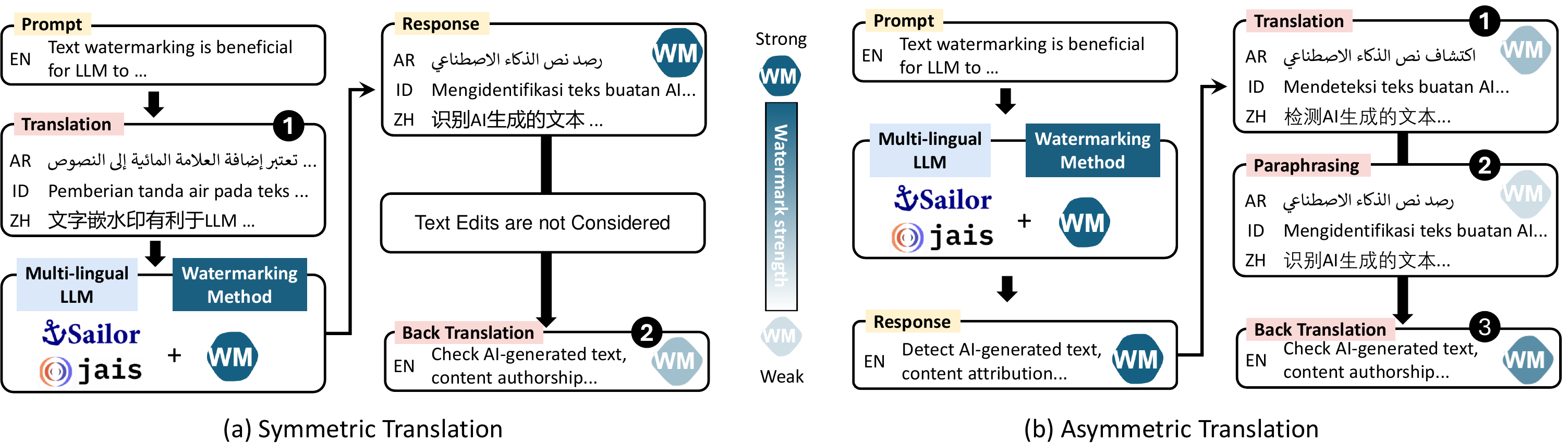}
  \caption{Existing symmetric cross-lingual attacks in comparison to our asymmetric attacks. (a) In Symmetrical translation attacks, a pivot language is used to \ding{182} translate the prompt. Then the original language is obtained \ding{183} without further edits are considered. (b) In Asymmetrical translation, the user \ding{182} translates to a target language. The user can use directly or \ding{183} optionally edit the text, further attenuating the watermark signal. Stage \ding{184} is also optional, which restores some of the watermark signal.}
  \label{fig:threat-model}
\end{figure*}

\section{Introduction}
The advancement of Large Language Models (LLMs) has significantly transformed text generation across various domains, producing outputs that closely mimic human writing. This advancement has raised concerns within the research community regarding potential misuses, including academic misconduct, the spread of disinformation, and the creation of synthetic training data~\citep{bender2021dangers, xue2022estas, zheng2024trojfsp, xue2023trojfair,lou2024cr}. In response to these challenges, watermarking methods have been developed to differentiate between human-written and AI-generated texts ~\citep{aaronson_my_2022,kirchenbauer2023watermark,kuditipudi2023robust,he2024can,dathathri2024scalable,chang2024postmark}.

Watermarking involves embedding a signal in AI-generated texts to identify the generating LLM using hypothesis testing. Specifically, watermarking offers theoretical guarantees regarding the detectability of the embedded signal by performing statistical inference on the generated text and testing against the null hypothesis. Since watermarking alters the original LLM output, it is crucial to ensure that the impact on text quality is minimal while maintaining the watermark's detectability. \footnote{Although steganography and watermarking share the practice of embedding signals, steganography primarily aims to conceal information through alterations of the text's meaning, whereas watermarking serves to assert text source without changing the text's semantics.}

Much of the existing literature on watermarking has primarily focused on the quality and detectability of watermarked texts, with a predominant emphasis on English texts. While these methods are theoretically language-agnostic, cross-lingual studies can reveal new adversarial scenarios in which watermark signal could be removed, and that have yet to be thoroughly investigated.

Existing studies on watermarking robustness have largely regarded the interplay between languages by back translation in which the English watermarked text is translated to some pivot language then by translating back to English, weakening the watermark signal ~\citep{kuditipudi2023robust, zhao2023provable, pang2024no, ghanim2024jailbreaking}. This overlooks other potential adversarial scenarios that may emerge following a translation to non-English languages.

A recent work by \citet{he2024can} attempted to explore adversarial scenarios within a cross-lingual context with specific translation attacks (e.g., Cross-lingual Watermark Removal Attack CWRA) to bypass watermarking by first obtaining a response from an LLM in a pivot language, which is then translated into the target prompt language. As can be shown in Figure~\ref{fig:threat-model}-a, we refer to this type of translation including back-translation as symmetrical, since an attacker does not utilize the pivot language. However, more practical adversarial scenarios are not studied, especially in the settings where users with cross-lingual knowledge might take to remove the watermark. As shown in Figure~\ref{fig:threat-model}-b, we call this asymmetrical since the user may use other languages without intentionally removing the watermark. This oversight accounts for both unintentional alterations made by users for clarity or context adjustment and sophisticated adversaries intentionally refining the text to destroy watermark traces such as CWRA. Additionally, a broader assessment of text quality under watermarking techniques in diverse languages is scarce but necessary, particularly given linguistic differences.

In this paper, we evaluate four representative watermarking methods under two high-level themes. First, syntactical watermarking, which involves syntax changes to the generated text by manipulating the logits before the decode stage. Second, Semantic watermarking, which involves semantics manipulation of the generated text before the decode stage as well. The syntactical methods we consider are KGW ~\citep{kirchenbauer2023watermark}, EXP ~\citep{aaronson_my_2022}, and Unigram ~\citep{zhao2023provable}. For the semantic methods, we choose XSIR ~\citep{he2024can} as it aligns with our cross-lingual investigation in this paper. Specifically, this paper addresses the following research questions.

\begin{itemize}[leftmargin=*, nosep, topsep=0pt, partopsep=0pt, parsep=5pt]
\item \textbf{RQ1}: How do watermarking methods perform across languages in terms of detectability, text quality, and diversity?
\item \textbf{RQ2:} How resilient are watermarking methods against existing cross-lingual attacks, particularly translation attacks?
\item \textbf{RQ3}: How do different adversarial approaches to cross-lingual watermark evasion perform? Specifically, how do attacks fare under our asymmetrical threat model, where the output language differs from the pivotal language?
\end{itemize}

Our investigation reveals that current watermarking methods face significant challenges in cross-lingual settings. Furthermore, we find that traditional text quality metrics may not adequately capture the diversity of cross-lingual watermarked text, necessitating new quality evaluation approaches. The main contributions of this paper are as follows:
\begin{itemize}[leftmargin=*, nosep, topsep=0pt, partopsep=0pt, parsep=5pt]
    \item We investigate the role of text watermarking across four different languages and four popular text watermarking methods, with a concentration on analyzing watermark detectability and quality.
    \item We propose a new text diversity metric that utilizes Self-BLEU to effectively evaluate the quality of watermarked text generated in a cross-lingual context.
    \item We evaluate the detectability of watermarks under practical attack pipeline consisting of translation, translation then paraphrase, and subsequent translation to the original language, and highlight the results on a robust cross-lingual method.
\end{itemize}

\section{Background and Related Work}
\textbf{Syntactical-based Watermarking.} Syntactical-based watermarking manipulates log-probabilities of generated text. KGW~\citep{kirchenbauer2023watermark, kirchenbauer2023reliability} and EXP~\citep{aaronson_my_2022} pioneered this by generating watermarks based on hashing previous tokens, introducing hypothesis testing with low False Positive Rate (FPR) for detection. Subsequent research expanded on these approaches: Unigram~\citep{zhao2023provable} used a pre-determined key for all generations; ~\citet{kuditipudi2023robust, christ2024undetectable} developed distortion-free watermarking to enhance the quality of watermarked text; ~\citet{dathathri2024scalable} introduced speculative sampling for generation at scale; and ~\citet{lu2024entropy, lee2023wrote} explored token entropy's role in watermark detectability.

\noindent\textbf{Semantic Watermarking.} To address paraphrasing and back-translation attacks that compromise syntactical watermarks, semantic approaches emerged. SIR~\citep{liu2024a} introduced semantic hashing of context rather than tokens. XSIR~\citep{he2024can} extended this with cross-lingual settings to counter translation attacks. ~\citet{chang2024postmark} developed a blackbox semantic watermarking for closed-source LLMs, while ~\citet{hou2023semstamp, hou2024k} introduced sentence-level clustering. \review{However, their rejection sampling approach slows watermarked text generation compared to methods in our benchmark}.

\noindent\textbf{Post-hoc Detection Approaches.}
Our evaluation focuses on proactive detection with embedded watermarks, but passive/discriminator methods exist for AI-generated text detection. ~\citet{tian_gptzero_2023,mitchell_detectgpt_2023,gehrmann2019gltr} use statistical patterns with discriminator models to differentiate human from AI written text. ~\citet{alshammari2024toward} detects AI-generated Arabic text using diacritics, while ~\citet{abdelnabi2021adversarial} embeds hidden watermarks by modifying transformer internals.

\noindent\textbf{Watermarking via Backdoors.} Recent work has explored backdoor-based approaches for fingerprinting and watermarking LLM outputs. \citet{xu2024instructional} used a technique to embed model-specific fingerprints through backdoor triggers in instruction tuning. Unlike traditional watermarking that modifies token probabilities during inference \cite{kirchenbauer2023watermark, aaronson_my_2022, kuditipudi2023robust}, backdoor approaches implant distinctive generation patterns during model training \citep{qi2021hidden,kurita2020weight} or at inference time \citep{al2023trojbits,xue2023trojllm,zheng2023trojvit}. This creates persistent model behaviors that can be detected with special inputs that trigger the backdoor.

\section{Threat Model}
We propose a threat model for text watermarking that addresses realistic removal scenarios. Our model includes various attack pipelines including translation, paraphrasing, and combinations thereof as illustrated in Figure \ref{fig:threat-model}. We focus on closed-source LLMs where model owners are the primary victims of watermark removal attacks.

\noindent\textbf{Attacker Knowledge:} We consider multilingual adversaries ranging from naive (unaware of watermarks) to advanced (deliberately trying to evade detection). Importantly, regular users may unintentionally remove watermarks through normal multilingual usage patterns.

\noindent\textbf{Attacker Capability:} Adversaries can access AI chat interfaces, translation APIs, and have the ability to paraphrase or edit text, potentially disrupting watermark signals.

\noindent\textbf{Real-world Scenarios:}
In an academic setting, non-English speaking professors providing English questions while allowing answers in native languages, enabling students to use AI-generated content with translation. This is not uncommon especially for science subjects where English is the language of the content is in English. Another scenario could take place in media and information sharing in which the post/content is translated from English to local languages and edited for specific audiences, inadvertently removing watermarks. We call these types of translations as asymmetric as the language of the utilized generation is different from that of the prompt.

These scenarios are common yet rarely considered in current watermark threat models. Robust watermarking systems must address multilingual use cases rather than focusing solely on using a language as pivotal to remove the watermark without utilizing it.

\section{Methodology}\label{sec:methods}
\subsection{\review{Benchmark Challenges}}
\noindent\review{\textbf{Quality Metrics.}} Evaluating watermarked text quality is essential for assessing watermark methods. Most studies focus on English, so we introduce a pipeline for multi-lingual LLMs. Initially, we used perplexity (PPL), but it wasn't effective across languages. PPL results can be shown in Appendix~\ref{sec:appendix-quality}, Figures~\ref{fig:z-scores-vs-ppls-kgw-unigram} and ~\ref{fig:z-scores-vs-ppls-others}. For better assessment, we use GPT-Judger from Singh et al. (2023) with an advanced OpenAI model. However, since we assess watermarking in multiple languages, evaluating quality with one method may not be sufficient to capture any possible linguistics nuances not captured in previous monolingual works. For example, multiple studies have shown biases introduced by the so called llm-as-a-judge paradigm in which a LLM is given multiple options to choose from~\citep{zheng2309large,pezeshkpour2023large,ye2024justice,koo2023benchmarking}. Therefore, we augment our quality evaluation with an evaluation of the judger itself by conducting fairness studies for any possible biases toward a specific language. Additionally, due to k-gram repetitions in watermarking, we apply Self-BLEU (SB) \citep{zhu2018texygen} to measure diversity and repetition. We combine the results from GPT-Judger and SB to create a metric that adjusts diversity more effectively, and provides insights into evaluating watermarked text quality in cross-lingual settings.

\noindent\review{\textbf{Text Interpolations in Cross-lingual Settings.}}
Existing literature on watermark attacks has primarily focused on intentional attacks designed to remove watermarks. In these studies, non-English languages are typically used as pivotal languages for back-and-forth translation to weaken or remove the watermark signal. However, this approach overlooks more practical and nuanced cross-lingual usage scenarios.
Watermarked text in multilingual settings faces challenges from translation and meaningful use in the intermediate languages. Adversaries with multilingual capabilities may utilize, edit, and distribute the content in various languages. These scenarios present more complex threats than simple pivotal translation attacks.

\noindent\textbf{Adjusted Diversity AD.} Because SB alone can be misleading in measuring the diversity of text --for example mixing tokens from different languages yields very low SB score, hence more diversity in text-- we create a new diversity metric that reflects unwanted diversity or excessive repetition in text by leveraging the Judger's coherency criterion scores as follow:
\begin{equation}\label{eq:AD-metric}
\small
    \text{AD} = w \times \text{SB} + 
    (1-w) \times (1 - \text{NC})
\end{equation}
 Where NC indicates Normalized Coherency score from GPT-Judger, $w$ is a weight between 0 and 1 that can be adjusted based on how much importance we want to give to each metric. The term $(1 - \text{NC})$ inverts the coherency score from GPT-Judger so that a low coherency score (indicating problematic text) contributes to a higher "unrealistic diversity" in the text. In our experiment, we choose $w$ to be $0.3$ as SB doesn't catch the semantic level of the text as GPT-Judger does, yet SB could give a subtle indication of repetition in the text.

\subsection{Watermark Methods}
\textbf{KGW \citep{kirchenbauer2023watermark}.} The KGW method embeds a watermark signal in generated text by manipulating log-probabilities (logits) of next token. The vocabulary $v$ is divided into green and red lists based on a split ratio $\gamma$. A secret key $S_k$ and a hash of the previous $k-1$ token ids seed a pseudorandom generator to produce the next token $k$. The generating model is either limited to the green list (hard watermark) or biased toward it (soft watermark) by adding a small value $\delta$ to the logits. Soft watermarking manages low-entropy contexts with few green list options. 

\noindent\textbf{Unigram \citep{zhao2023provable}.} Unigram, like KGW, divides the vocabulary $v$ into green list $v_g$ and red list $v_r$. The key distinction is that in Unigram, these lists are consistent for all generated tokens over the course of the generation process, as it does not utilize hashing with previous token IDs. Instead, Unigram employs the sha256 hashing algorithm, using a secret key $S_k$ to partition the vocabulary into the green and red lists. Subsequently, it applies the soft-watermark technique to adjust the logit values. This approach is expected to reduce the effectiveness of watermark frequency counting removal attacks, since the hashing process does not incorporate previous token IDs. Both KGW and Unigram use the following equation to detect the watermark.
\begin{equation}
\small
    z = (|s|_G - \gamma |s|) / \sqrt{|s|\gamma(1-\gamma)}
\end{equation}
where $|s|_G$ is the number of green tokens in the generated text, and $\gamma=\frac{|v_G|}{|v|}$.

\noindent\textbf{EXP \cite{aaronson_my_2022}.} EXP uses exponential minimum sampling, which is a variant of the Gumbel trick ~\citet{papandreou2011perturb}, to bias the distribution of the next token generation. Specifically, like KGW, EXP uses $k-1$ context-window to seed a pseudorandom generator (PRG) to generate the next token $k$. However, instead of using soft watermark or applying a $\delta$ value to the logits, EXP uses PRG to generate random numbers $r_{t,i}$, which is the same size as the vocabulary of the generating model. Then, at position $t$, token $i$ is sampled by maximizing the following quantity. 
\[
\small
    \text{argmax}_i(r_{t,i}^{1/p_{t,i}})
\]
when $p_{t,i}$ is very small, token $i$ will only be chosen if $r_{t,i}$ is close to one, which is very unlikely to happen. In terms of randomness, this sampling method will return the same token every time the same $k-1$ context is used for the PRG. The watermark is then detected by the following equation.
\begin{equation}
\small
    \sum_{t=1}^n log(\frac{1}{1-r_{t,i}})
\end{equation}

\noindent\textbf{XSIR \cite{he2024can}.} XSIR is designed to enhance cross-lingual watermarking by ensuring that semantically similar prefix texts receive similar logit biases. Instead of directly hashing token IDs, XSIR hashes a semantic chunk of the prefix text to generate the logit bias for the next token \( k \). This approach ensures that different prefix texts with similar meanings produce comparable logit bias distributions:

\begin{equation}
\small
    \text{Sim}(\Delta(x), \Delta(y)) \approx \text{Sim}(E(x),E(y))
\end{equation}
where \( E \) is a multilingual embedding model, and \( \Delta \) is the function that determines the watermark logit bias distribution for all tokens in the vocabulary. XSIR uses semantic clustering to assign consistent biases, ensuring tokens with similar prefixes or context receive similar watermarking bias.

Besides, XSIR also enforces consistent biases across words within the same semantic cluster. Specifically, words that share the same meaning across different languages are assigned identical biases:

\begin{equation}
\small
    C(i) = C(j) \Rightarrow \Delta_{C(i)} = \Delta_{C(j)}
\end{equation}
where \( C(i) \) represents the cluster index of word \( i \). As a result, if tokens \( i \) and \( j \) are semantically equivalent, they receive the same logit bias, preserving watermark consistency across translations.


\begin{figure*}[h!]
\centering
\includegraphics[width=\linewidth]{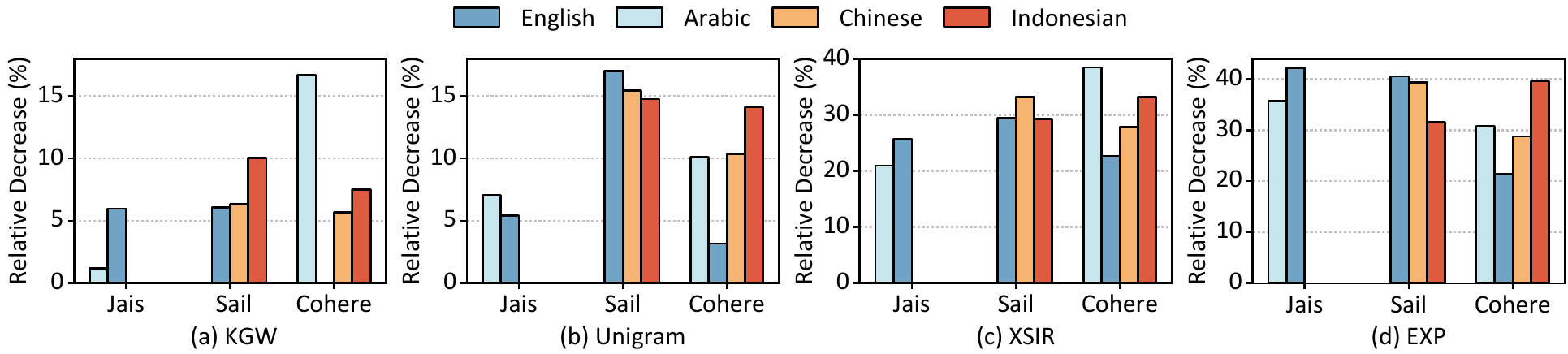}
\caption{GPT-judge coherency criterion results. We compute the average of watermarked and unwatermarked scores for $500$ generations. We use Jais model for Arabic and Sail for Chinese and Indonesian. We use Cohere to represent all languages.}
\label{fig:gptjudge-results}
\end{figure*}

\section{Experimental Setup}
\noindent\textbf{Choice of Models.} Due to scarcity of open-source LLMs that universally support our targeted languages, we employ different models suitable for different languages. For Chinese and Indonesian, we utilize the Sailor2 1B and 8B variants~\cite{sailor2report}. For Arabic text generation, we use the Jais family 6.7B model~\cite{jaisfamilymodelcard}, while perplexity is assessed using Acegpt 7B~\cite{huang2023acegpt}. For English, we generate text using both aforementioned models. The use of varying models across languages is essential to accommodate the lack of universally compatible models but ensures credible cross-lingual watermarking assessments. Recently, a model that support all our targeted languages was released ~\citep{cohere2025commandaenterprisereadylarge} in which all of our targeted languages are supported\footnote{\url{https://huggingface.co/CohereLabs/c4ai-command-r7b-12-2024}}. We use this model to add experiments for an instruction-following and text-completion task with more languages.

\noindent\textbf{Dataset Selection.} For all experiments, we use $500$ examples from the C4 dataset~\cite{JMLR:v21:20-074}, which is available in all evaluated languages. For English, we follow previous work by using the RealNewsLike split; for other languages, we use the main training split, as RealNewsLike is available only for English\footnote{\url{https://huggingface.co}}. We also utilized a cleaned version of LFQA dataset from this work ~\cite{krishna2023paraphrasing}. We use this dataset along with instruction-following prompting for information entropy analysis in different languages to further assess the quality of text with watermarking.

\noindent\textbf{Watermarking Parameters.} We evaluate watermarking across four methods: KGW, Unigram, XSIR, and EXP. For KGW and Unigram, parameters include green list ratios $\gamma \in {0.1, 0.5, 0.9}$ and bias values $\delta \in {2, 5, 10}$. A context size of 1 is applied for KGW’s \emph{lefthash}. For EXP, a context length of 4 and a $p$-value threshold of $10^{-4}$ are used. XSIR divides the vocabulary with $\gamma=0.5$ and bias values $\delta \in {2, 5, 10}$. Standard settings of $\gamma=0.5$ and $\delta=2.0$ are typically used unless otherwise noted, based on optimal empirical results. Additional parameter results are detailed in the appendices.

\noindent\textbf{Watermarking Metrics} Following prior studies~\cite{kirchenbauer2023watermark,zhao2023provable,dathathri2024scalable}, watermark detection is analyzed via ROC curves, focusing on the trade-off between the false and true positive/negative rates. An empirical threshold is determined using a comparison between unwatermarked and watermarked text scores to enhance multilingual watermark consistency. See Appendix~\ref{sec:appendix-detection} for detailed performance metrics using specific threshold values.

\noindent\review{\textbf{Language Bias Assessment.}} We conducted controlled experiments to identify potential language biases in LLM judges. Starting with English examples from the C4 dataset, we used GPT-3.5-Turbo to create perfect translations in seven languages: English, Arabic, Chinese, Indonesian, Persian (Farsi), German, and Japanese. This approach established ground truth by ensuring identical content across all languages, allowing us to isolate language-specific biases. For methodological rigor, we randomized text presentation order and conducted multiple runs with different seeds. See Appendix~\ref{sec:appendix-gpt-fairness} for complete experimental details.

\noindent\textbf{Execution Details.} Experiments involve generating watermarked text in various languages, with processes taking approximately 10 GPU hours for KGW and Unigram, 7 hours for XSIR, and 3 hours for EXP. For qualitative assessment, the GPT-4o-mini-2024-07-18 serves as the GPT-based judger, conducting quality evaluations in roughly 20 minutes for 500 generations. Since translations are heavily used in this work, we opt for the open-sourced and easy to use OPUS-MT models~\cite{tiedemann2020opus,tiedemann2022democratizing} which are used by previous works such as~\citet{kuditipudi2023robust}.

\section{Results}




\subsection{RQ1: Watermarking Performance under Cross-lingual Setting}
\noindent\textbf{Quality.} We evaluate text quality through two complementary approaches: GPT-Judger and our proposed diversity metrics. This dual evaluation provides a comprehensive view of how watermarking affects text quality across languages.

\begin{figure}
\centering
\includegraphics[width=0.9\linewidth]{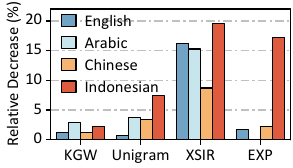}
\caption{GPT-judge coherency criterion results for Cohere model on LFQA dataset, computed as the average of watermarked and unwatermarked scores for $500$ generations.}
\vspace{-0.1in}
\label{fig:gptjudge-results-lfqa-only}
\end{figure}

In Figures~\ref{fig:gptjudge-results} and ~\ref{fig:gptjudge-results-lfqa-only} we show the result of obtaining the decrease percentage in quality after watermarking. In these figures, we only present the coherency criterion since its scores reflect the most affected criterion in all languages according to GPT-Judger. The relative decrease in the figure is calculated as:
\begin{equation}
    \small
    (\frac{\text{unwatermarked\_score}-\text{watermarked\_score}}{\text{unwatermarked\_score}})\times 100
\end{equation}
We use C4 to reflect a more generalized text completion scenario, and we use LFQA for low-entropy instruction-following completions. The relative impact of coherency decrease varies between languages in the same watermarking method, particularly for C4 dataset. While Chinese and Indonesian often show higher coherency decrease than English and Arabic in several methods, this impact is reduced in low-entropy completions. While this suggests easier watermarking for instruction-following scenarios, it can affect the detectability of the watermark as shown in Table~\ref{tab:all-error-rates-lfqa} in the appendices. Among non-English languages, Arabic seems to be easier to watermark under KGW and Unigram methods, but more difficult to watermark under XSIR. In terms of watermarking method, KGW has minimal impact of quality for both datasets, while XSIR and EXP showcase the largest degrade of the quality across all languages. In Appendix ~\ref{sec:appendix-quality}, Figure~\ref{fig:gptjudge-results-more-langs}, we show more results for additional languages.

\begin{table}[h]
    \centering
    \small
    \setlength{\tabcolsep}{4pt}
    \renewcommand\arraystretch{0.9}
    \caption{GPT-Judger Final Verdict Analysis. A soft-win is recorded when the watermarked text is judged to be of equal (Tie) or superior quality (Hard-win) compared to the non-watermarked version, reflecting the method's ability to preserve text quality. Higher soft-win rates indicate better quality retention after watermarking.}
    \begin{tabular}{lcccc}\toprule
        Method & Language & Hard-Win \(\uparrow\) & Tie \(\uparrow\) & Soft-Win \(\uparrow\) \\
        \midrule
        \multirow{4}{*}{KGW} & English & 0.42 & 0.05 & 0.47 \\
        & Arabic & 0.41 & 0.15 & 0.56\\
        & Chinese & 0.31 & 0.26 & 0.57\\
        & Indonesian & 0.29 & 0.18 & 0.47\\
        \midrule
        \multirow{4}{*}{Unigram} & English & 0.38 & 0.08 & 0.46 \\
        & Arabic & 0.37 & 0.12 & 0.49\\
        & Chinese & 0.24 & 0.27 & 0.51\\
        & Indonesian & 0.27 & 0.14 & 0.41\\
        \midrule
        \multirow{4}{*}{XSIR} & English & 0.22 & 0.12 & 0.34 \\
        & Arabic & 0.20 & 0.06 & 0.26\\
        & Chinese & 0.11 & 0.24 & 0.35\\
        & Indonesian & 0.16 & 0.12 & 0.28\\
        \midrule
        \multirow{4}{*}{EXP} & English & 0.10 & 0.04 & 0.14\\
        & Arabic & 0.14 & 0.10 & 0.25\\
        & Chinese & 0.07 & 0.24 & 0.31\\
        & Indonesian & 0.15 & 0.12 & 0.26\\
        \bottomrule
    \end{tabular}
    
    \label{tab:soft-win-rates}
\end{table}

To further analyze the GPT-Judger results, we calculate soft-win rates in Table~\ref{tab:soft-win-rates}. Among all methods, KGW achieves the highest soft-win rate. Nonetheless, the GPT-Judger occasionally struggles to deliver definitive judgments, particularly for non-English texts, as indicated by the high tie rates.

\noindent\review{\textbf{LLM Judge Language Bias.}} Our fairness experiments reveal significant language biases in GPT-Judger's final verdict, which may explain some of the result in Figure~\ref{fig:gptjudge-results}. The results of these experiments are presented in Appendix ~\ref{sec:appendix-gpt-fairness} Tables ~\ref{tab:trans-verdict-percentages} and ~\ref{tab:para-verdict-percentages}. When evaluating identical content translated into different languages (Translation Experiment), we found clear preferences for certain languages regardless of content quality. German was most preferred, followed by Arabic and English, while \emph{Chinese}, \emph{Persian}, and \emph{Indonesian} received consistently lower scores. In a complementary "Paraphrase Experiment" using same-language text pairs, we observed judges showing higher TIE rates for some languages over the others. Additionally, the judger consistently favored paraphrased texts generated by their own model family over original texts. These biases suggest that the high TIE rates in our watermarking evaluations likely affected by the biased preferences we saw in the "translation experiment" rather than actual quality equivalence judgments. For detailed analysis including position bias investigation and cross-language comparison metrics, see Appendix~\ref{sec:appendix-gpt-fairness}. These findings align with token bias studies by \citet{zheng2309large} and model self-preference observations by \citet{ye2024justice}, and therefore, \emph{llm-as-a-judge paradigms should be coupled with some sort of debiasing procedures to ensure fair choice between texts of different languages}.

\begin{table*}[ht]
    \renewcommand\arraystretch{0.9}
    \centering
    \small
    \caption{SelfBleu results for all watermark methods. For KGW, Unigram and XSIR, $\gamma=0.5$ and $\delta=2.0$. We consider this setting is the best for these methods with different languages. More results with varying the $\gamma$ and $\delta$ values in are in Appendix.}
    \setlength{\tabcolsep}{12pt}
    \begin{tabular}{cccccc}
        \toprule
        \multirow{2}{*}{Method} & \multirow{2}{*}{Language} & \multicolumn{2}{c}{self-bleu ($\downarrow$ more diverse)} & \multicolumn{2}{c}{AD ($\downarrow$ better)}\\
        \cmidrule(lr){3-4}\cmidrule(lr){5-6}
        & & watermarked & unwatermarked & watermarked & unwatermarked\\
        \midrule
        \multirow{4}{*}{KGW} & English & 0.16 & 0.16 & 0.38 & 0.34\\
        & Arabic & 0.11 & 0.12 & 0.36 & 0.35\\
        & Chinese & 0.04 & 0.04 & \textbf{0.44} & 0.40\\
        & Indonesian & 0.10 & 0.10 & \textbf{0.43} & 0.38\\
        \midrule
        \multirow{4}{*}{Unigram} & English & 0.23 & 0.17 & 0.40 & 0.35\\
        & Arabic & 0.16 & 0.12 & 0.41 & 0.35\\
        & Chinese & 0.02 & 0.04 & \textbf{0.47} & 0.40\\
        & Indonesian & 0.12 & 0.10 & \textbf{0.46} & 0.37\\
        \midrule
        \multirow{4}{*}{XSIR} & English & 0.20 & 0.17  & \textbf{0.49} & 0.32\\
        & Arabic & 0.14 & 0.12 & \textbf{0.45} & 0.33\\
        & Chinese & 0.03 & 0.04 & \textbf{0.55} & 0.38\\
        & Indonesian & 0.12 & 0.11 & \textbf{0.53} & 0.37\\
        \midrule
        \multirow{4}{*}{EXP} & English & 0.19 & 0.17 & \textbf{0.57} & 0.29 \\
        & Arabic & 0.20 & 0.12 & \textbf{0.55} & 0.31\\
        & Chinese & 0.13 & 0.04 & \textbf{0.61} & 0.38\\
        & Indonesian & 0.21 & 0.10 & \textbf{0.57} & 0.36\\
        \bottomrule
    \end{tabular}

    \label{tab:self-bleu}
\end{table*}

\begin{figure*}[th]
\subfigure{\includegraphics[width=0.98\columnwidth]{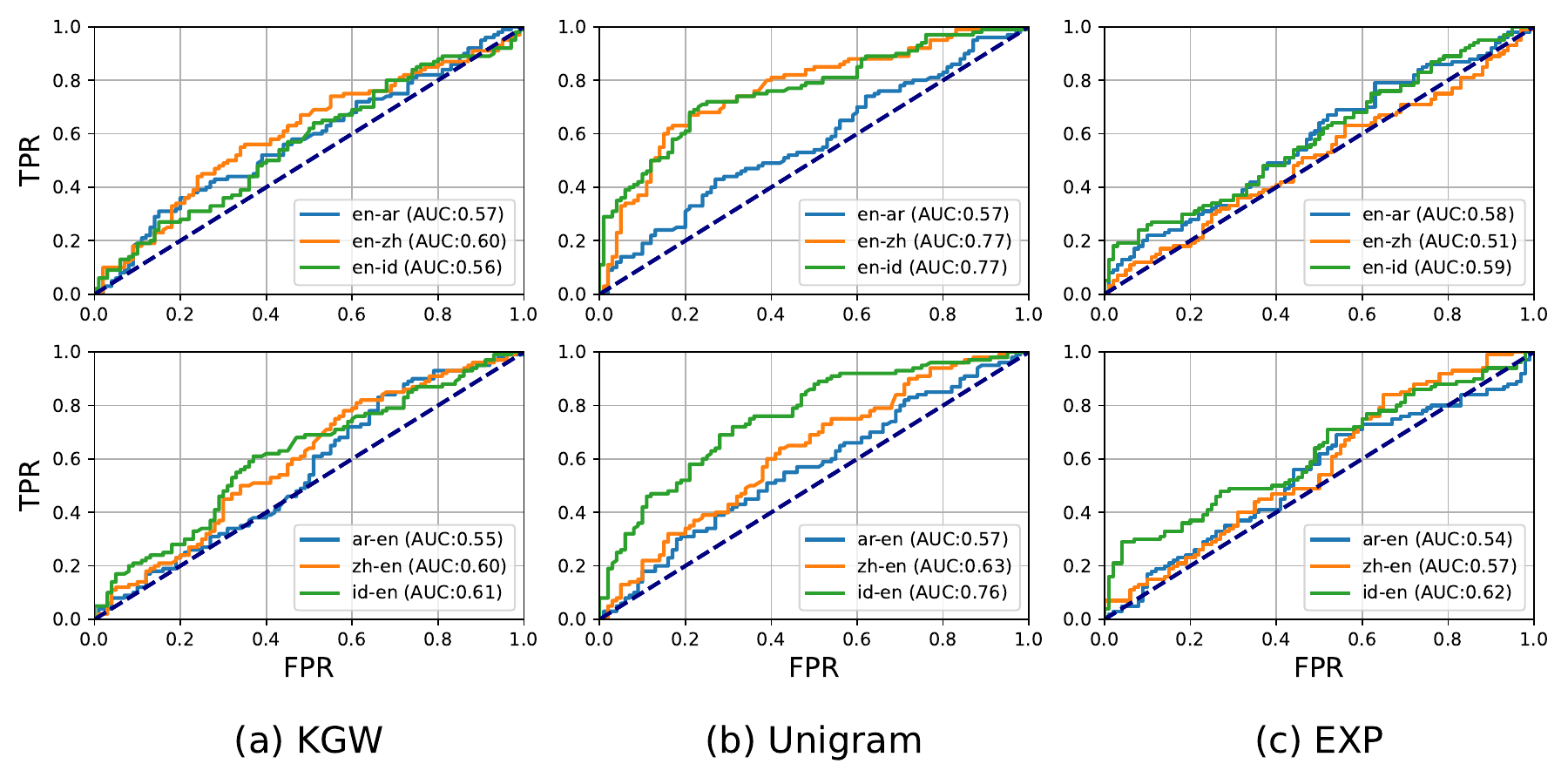}\label{fig:roc-curve-kgw-unigram-exp-translations}}
\hfill
 \subfigure{\includegraphics[width=0.98\columnwidth]{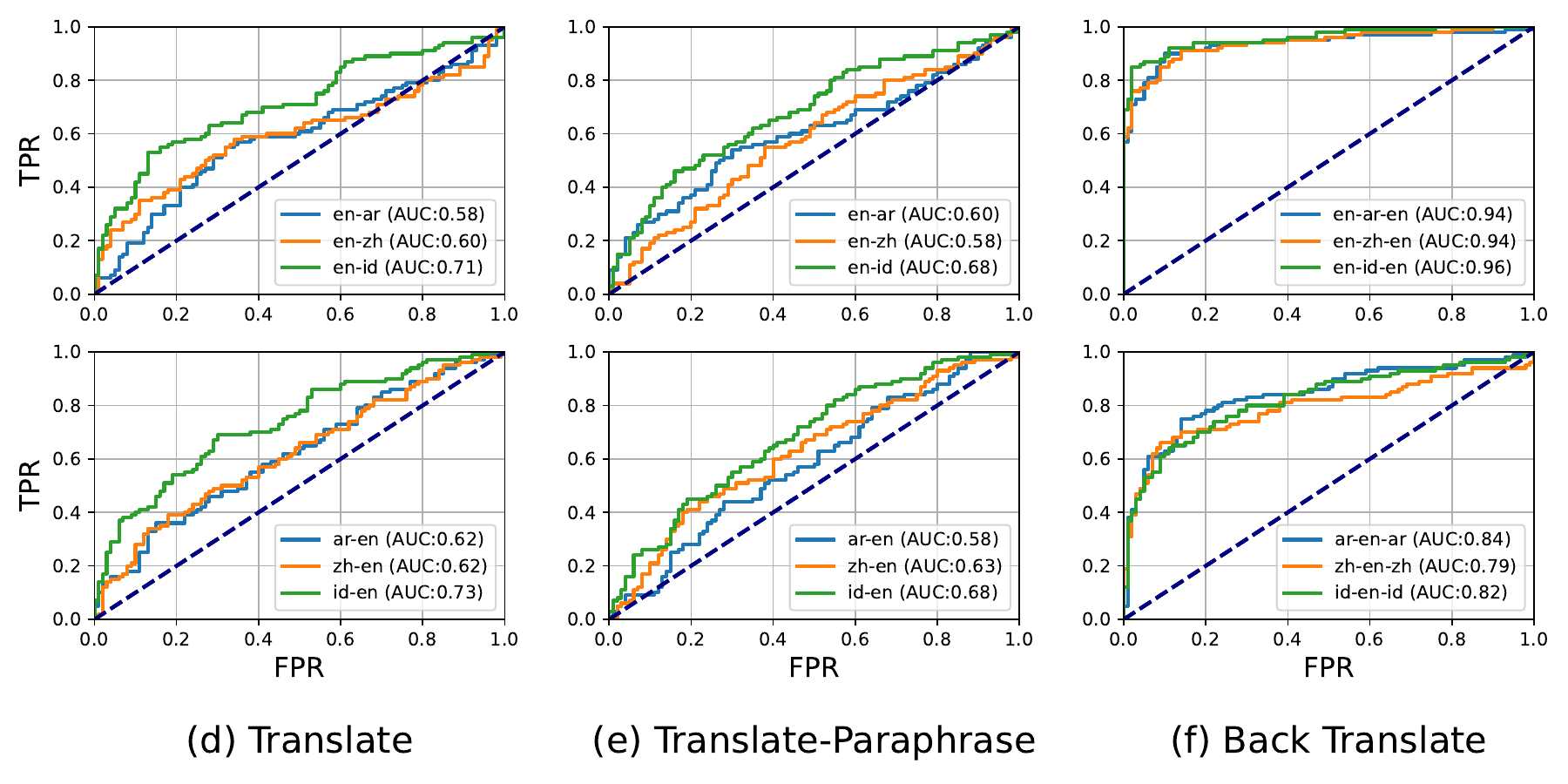}\label{fig:roc-curve-xsir-all-attacks}}
  \vspace{-3mm}
  \caption {Watermark detection ROC curves with AUC after attacks. For all attacks, the watermark threshold is calculated automatically by comparing unwatermarked and attacked watermarked text score for $100$ generations. \textbf{Left:} we perform symmetrical translation attacks on KGW, Unigram, and EXP. \textbf{Right:} Only the leftmost column represent symmetrical translation attack setting. We further apply more invasive attacks to XSIR to highlight its robustness.}
  \label{fig:roc-curves-all-methods-after}
  \vspace{-3mm} 
\end{figure*}

\noindent\textbf{Diversity.}
To more objectively quantify quality of text with a predefined text quality metric, we introduce our new metric rooted in the Self-BLEU (SB) metric. In our experiments, the Adjusted Diversity (AD) is used to further assess the quality of watermarked text along with the assessment provided by the GPT-Judger's coherency scores. Previous research has used the SB metric to assess the diversity of watermarked text \citep{dathathri2024scalable}. Table~\ref{tab:self-bleu} includes results from SB and the Adjusted Diversity (AD) metrics for various watermark methods. While SB suggests high diversity for Chinese and Indonesian, it can be misleading; low SB scores (e.g., $0.04$ for Chinese) might indicate text degradation, as shown by higher AD scores ($\geq 0.44$). This pattern holds across other languages. When we employ larger $\delta$ values with $\gamma=0.5$, AD ratios are higher compared to SB, indicating unrealistic diversity in watermarked text. More detailed results with different hyper-parameters are in Table~\ref{tab:self-bleu-more} in Appendix~\ref{sec:appendix-quality}.

\noindent\textbf{Detectability.}
Our detection analysis examines both performance without attack and attack resilience across languages. Prior to any attacks, all methods demonstrate strong detectability as shown in Figure~\ref{fig:roc-curves-all-methods-before} in Appendix~\ref{sec:appendix-detection} in which all methods across all languages achieve $\geq0.99$ AUC scores. For KGW, Unigram, and XSIR, we identified $\delta=2.0$ and $\gamma=0.5$ as the optimal parameters yielding the best detection results. However, the role of employing different hyper-parameters for detection differs from that for text quality. When larger values of $\gamma$ are used with $delta=2.0$, detection scores are adversely affected, especially for Unigram method. We show detailed results in Figure~\ref{fig:roc-curves-kgw-unigram-before-zoomed} in which we illustrate the close relationship between $\gamma$ and $\delta$ for methods like KGW and Unigram in Appendix~\ref{sec:appendix-detection}.

\subsection{RQ2: Watermarking Resilience to Translation Attacks.} 
Regarding attack resilience, all methods are vulnerable to our attack pipeline. In Figure~\ref{fig:roc-curves-all-methods-after}~(left), we perform \textit{asymmetrical} translation attacks against KGW, Unigram, and EXP. The results showcase the watermark detectability under translation attacks varies by language. For instance, KGW shows the lowest AUC of $0.55$ for Arabic-English, while Unigram presents a worst-case AUC of $0.57$ for English-Arabic and Arabic-English. For EXP ( Figure~\ref{fig:roc-curves-all-methods-after}~(c)), all attacks are notably invasive, likely due to its larger $k$-gram hashing window of $4$ compared to KGW's recommended window of $1$.

In Figure~\ref{fig:roc-curves-all-methods-after}~(right), we evaluate all of our \textit{asymmetrical} attacks to the more cross-lingual XSIR method. Although XSIR was developed for \textit{symmetrical} attacks, it remains susceptible to \textit{asymmetrical} attacks. This vulnerability underscores the severity of attack scenarios where the pivot language differs from the output language. XSIR's semantic clusters vary across languages, weakening its watermark robustness when the pivot and output languages differ. In contrast, back-translate attacks preserve the watermark (Figure~\ref{fig:roc-curves-all-methods-after}~(f)).

Finally, KGW and EXP have more predictable results than XSIR and Unigram where different languages affect the watermark signal differently. For instance, detection outcomes for translation and translation-paraphrase attacks remain consistently tight across languages for KGW and EXP, whereas XSIR and Unigram exhibit greater variability, with some languages deviating significantly. This behavior may stem from technical challenges such as preserving text length during translation.

\subsection{RQ3: Asymmetrical Translation Attacks Raise a New Threat}
Despite XSIR being specifically designed for cross-lingual scenarios, our results shown in Figure~\ref{fig:roc-curve-xsir-all-attacks} demonstrate that it remains vulnerable to post-generation translation attacks. Particularly, the vulnerability is pronounced for Chinese and Arabic languages, suggesting that XSIR's effectiveness varies significantly by language. This contrasts with XSIR proposed adversarial scenario of translating prompts before generation and translating them back into the original language.

From Figures ~\ref{fig:roc-curves-all-methods-after} and ~\ref{fig:roc-curves-syntactical-methods-after}, our analysis reveals a critical asymmetry: starting the prompt with \textit{ENGLISH} seems to yield better detection rates than starting with other languages. In other words, \emph{considering the interpolation of languages in the context of only using a language as pivotal to evade watermarking is not enough to assess the robustness of watermarks in cross-lingual manner}.

The attacks presented are invasive, significantly weakening the watermark signal, but can be countered through various strategies. We believe that developing more resilient cross-lingual watermarking techniques should integrate KGW and EXP predictability with the robust Unigram method. XSIR could be instrumental here, as it manipulates text semantics over syntax.

Our investigation into XSIR revealed that its shortcomings in handling translation attacks are largely technical. XSIR clusters semantically similar words across languages by constructing a graph from a comprehensive dictionary that contains pairs of related words in multiple languages. It then creates connected components (CCs) to connect all similar nodes together through string matching. This creates a mapping that is used to broadcast the same bias to the logits of all related words in a cluster. However, the process in which the words in the dictionary are connected to a specific model's vocabulary is tokenizer-dependent. This is because some tokenizers store vocabularies as unicode characters as a result of the encoding process of tokenization methods of different vocabularies. English tokens are usually stored as English alphabets, but this is not the case for non-English tokens. This creates clusters of matching English tokens only, bypassing similar words in other languages.

To enhance XSIR's robustness, one can modify the detection process to translate a generated text to the original language of the prompt. Given that English handles XSIR cluster effectively, translating non-English outputs to English before detection could ensure watermark signal identification. However, this only works if the language in which the first prompt was generated is English. Clearly, this is not a viable option in a cross-lingual setting in which the prompt language can be non-English.

\section{Conclusion}
Our study reveals significant variations in LLM watermarking effectiveness across languages. KGW best preserves text quality, while all methods show vulnerability to translation attacks, especially asymmetrical ones where pivot and output languages differ. We identified substantial language biases in LLM-based evaluations, with GPT-Judger showing clear preferences for certain languages over others. Our findings demonstrate that current watermarking approaches inadequately address cross-lingual complexities, as effectiveness depends not only on the target language but also on specific language paths during attacks. This highlights the need for more robust cross-lingual watermarking methods that maintain effectiveness across diverse linguistic contexts.

\section{Limitations}
In this study, we evaluate four distinct watermarking methods across four languages—English, Arabic, Chinese, and Indonesian—focusing on practical quality evaluation and removal attacks in a cross-lingual context. We also extend the investigation to Turkish and Hindi using the CohereAI model. However, we believe that more research should include a wider variety of languages - both low-resource languages and those with complex syntactic and grammatical features. Additionally, we have included studies on C4 and LFQA datasets but we think expanding the datasets beyond these two could be beneficial in ensuring final insights about a language or a watermarking method.

\section{Ethics Consideration}
Our research reveals critical vulnerabilities in text watermarking when subjected to cross-lingual translation attacks—specifically, the risk that the original language of a text can be concealed, allowing adversaries to evade detection and potentially disseminate harmful content. We acknowledge that such findings may be exploited by malicious actors, thereby posing a serious risk to digital authenticity and safety. However, the primary goal of our work is to illuminate these weaknesses so that more robust watermarking strategies can be developed and integrated into language models. In the interim, we propose simple yet effective countermeasures that can be readily incorporated by AI service providers. Our study employs publicly available data and models and is intended solely for academic research and the improvement of digital security. We strongly advocate for responsible disclosure and the continuous refinement of safeguards to ensure that AI technologies are deployed safely and ethically.

\section*{Acknowledgments}
We would like to thank the reviewers for their helpful comments that strengthened this paper. We extend our thanks to King Khalid University for its support during the lifetime of this project.

\bibliography{acl}

\begin{thebibliography}{46}
\providecommand{\natexlab}[1]{#1}

\bibitem[{Aaronson(2022)}]{aaronson_my_2022}
Scott Aaronson. 2022.
\newblock \href {https://scottaaronson.blog/?p=6823} {My {{AI Safety Lecture}} for {{UT Effective Altruism}}}.

\bibitem[{Abdelnabi and Fritz(2021)}]{abdelnabi2021adversarial}
Sahar Abdelnabi and Mario Fritz. 2021.
\newblock Adversarial watermarking transformer: Towards tracing text provenance with data hiding.
\newblock In \emph{2021 IEEE Symposium on Security and Privacy (SP)}, pages 121--140. IEEE.

\bibitem[{Al~Ghanim et~al.(2023)Al~Ghanim, Santriaji, Lou, and Solihin}]{al2023trojbits}
Mansour Al~Ghanim, Muhammad Santriaji, Qian Lou, and Yan Solihin. 2023.
\newblock Trojbits: A hardware aware inference-time attack on transformer-based language models.
\newblock In \emph{ECAI 2023}, pages 60--68. IOS Press.

\bibitem[{Alshammari and Elleithy(2024)}]{alshammari2024toward}
Hamed Alshammari and Khaled Elleithy. 2024.
\newblock Toward robust arabic ai-generated text detection: Tackling diacritics challenges.
\newblock \emph{Information}, 15(7):419.

\bibitem[{Bender et~al.(2021)Bender, Gebru, McMillan-Major, and Shmitchell}]{bender2021dangers}
Emily~M Bender, Timnit Gebru, Angelina McMillan-Major, and Shmargaret Shmitchell. 2021.
\newblock On the dangers of stochastic parrots: Can language models be too big?
\newblock In \emph{Proceedings of the 2021 ACM conference on fairness, accountability, and transparency}, pages 610--623.

\bibitem[{Chang et~al.(2024)Chang, Krishna, Houmansadr, Wieting, and Iyyer}]{chang2024postmark}
Yapei Chang, Kalpesh Krishna, Amir Houmansadr, John Wieting, and Mohit Iyyer. 2024.
\newblock Postmark: A robust blackbox watermark for large language models.
\newblock \emph{arXiv preprint arXiv:2406.14517}.

\bibitem[{Christ et~al.(2024)Christ, Gunn, and Zamir}]{christ2024undetectable}
Miranda Christ, Sam Gunn, and Or~Zamir. 2024.
\newblock Undetectable watermarks for language models.
\newblock In \emph{The Thirty Seventh Annual Conference on Learning Theory}, pages 1125--1139. PMLR.

\bibitem[{Cohere et~al.(2025)Cohere, Ahmadian, Ahmed, Alammar, Alizadeh, Alnumay, Althammer, Arkhangorodsky, Aryabumi, Aumiller et~al.}]{cohere2025commandaenterprisereadylarge}
Team Cohere, Arash Ahmadian, Marwan Ahmed, Jay Alammar, Milad Alizadeh, Yazeed Alnumay, Sophia Althammer, Arkady Arkhangorodsky, Viraat Aryabumi, Dennis Aumiller, et~al. 2025.
\newblock Command a: An enterprise-ready large language model.
\newblock \emph{arXiv preprint arXiv:2504.00698}.

\bibitem[{Dathathri et~al.(2024)Dathathri, See, Ghaisas, Huang, McAdam, Welbl, Bachani, Kaskasoli, Stanforth, Matejovicova et~al.}]{dathathri2024scalable}
Sumanth Dathathri, Abigail See, Sumedh Ghaisas, Po-Sen Huang, Rob McAdam, Johannes Welbl, Vandana Bachani, Alex Kaskasoli, Robert Stanforth, Tatiana Matejovicova, et~al. 2024.
\newblock Scalable watermarking for identifying large language model outputs.
\newblock \emph{Nature}, 634(8035):818--823.

\bibitem[{Gehrmann et~al.(2019)Gehrmann, Strobelt, and Rush}]{gehrmann2019gltr}
Sebastian Gehrmann, Hendrik Strobelt, and Alexander~M Rush. 2019.
\newblock Gltr: Statistical detection and visualization of generated text.
\newblock \emph{arXiv preprint arXiv:1906.04043}.

\bibitem[{Ghanim et~al.(2024)Ghanim, Almohaimeed, Zheng, Solihin, and Lou}]{ghanim2024jailbreaking}
Mansour Ghanim, Saleh Almohaimeed, Mengxin Zheng, Yan Solihin, and Qian Lou. 2024.
\newblock Jailbreaking llms with arabic transliteration and arabizi.
\newblock In \emph{Proceedings of the 2024 Conference on Empirical Methods in Natural Language Processing}, pages 18584--18600.

\bibitem[{He et~al.(2024)He, Zhou, Hao, Liu, Wang, Tu, Zhang, and Wang}]{he2024can}
Zhiwei He, Binglin Zhou, Hongkun Hao, Aiwei Liu, Xing Wang, Zhaopeng Tu, Zhuosheng Zhang, and Rui Wang. 2024.
\newblock Can watermarks survive translation? on the cross-lingual consistency of text watermark for large language models.
\newblock \emph{arXiv preprint arXiv:2402.14007}.

\bibitem[{Hou et~al.(2023)Hou, Zhang, He, Wang, Chuang, Wang, Shen, Van~Durme, Khashabi, and Tsvetkov}]{hou2023semstamp}
Abe~Bohan Hou, Jingyu Zhang, Tianxing He, Yichen Wang, Yung-Sung Chuang, Hongwei Wang, Lingfeng Shen, Benjamin Van~Durme, Daniel Khashabi, and Yulia Tsvetkov. 2023.
\newblock Semstamp: A semantic watermark with paraphrastic robustness for text generation.
\newblock \emph{arXiv preprint arXiv:2310.03991}.

\bibitem[{Hou et~al.(2024)Hou, Zhang, Wang, Khashabi, and He}]{hou2024k}
Abe~Bohan Hou, Jingyu Zhang, Yichen Wang, Daniel Khashabi, and Tianxing He. 2024.
\newblock k-semstamp: A clustering-based semantic watermark for detection of machine-generated text.
\newblock \emph{arXiv preprint arXiv:2402.11399}.

\bibitem[{Huang et~al.(2023)Huang, Yu, Zhu, Sun, Cheng, Song, Chen, Alharthi, An, Liu, Zhang, Chen, Li, Wang, Zhang, Sun, Wan, Li, and Xu}]{huang2023acegpt}
Huang Huang, Fei Yu, Jianqing Zhu, Xuening Sun, Hao Cheng, Dingjie Song, Zhihong Chen, Abdulmohsen Alharthi, Bang An, Ziche Liu, Zhiyi Zhang, Junying Chen, Jianquan Li, Benyou Wang, Lian Zhang, Ruoyu Sun, Xiang Wan, Haizhou Li, and Jinchao Xu. 2023.
\newblock \href {https://arxiv.org/abs/2309.12053} {Acegpt, localizing large language models in arabic}.
\newblock \emph{Preprint}, arXiv:2309.12053.

\bibitem[{Inception(2024)}]{jaisfamilymodelcard}
Inception. 2024.
\newblock \href {https://huggingface.co/inceptionai/jais-family-30b-16k-chat/blob/main/README.md} {Jais family model card}.

\bibitem[{Kirchenbauer et~al.(2023{\natexlab{a}})Kirchenbauer, Geiping, Wen, Katz, Miers, and Goldstein}]{kirchenbauer2023watermark}
John Kirchenbauer, Jonas Geiping, Yuxin Wen, Jonathan Katz, Ian Miers, and Tom Goldstein. 2023{\natexlab{a}}.
\newblock A watermark for large language models.
\newblock In \emph{International Conference on Machine Learning}, pages 17061--17084. PMLR.

\bibitem[{Kirchenbauer et~al.(2023{\natexlab{b}})Kirchenbauer, Geiping, Wen, Shu, Saifullah, Kong, Fernando, Saha, Goldblum, and Goldstein}]{kirchenbauer2023reliability}
John Kirchenbauer, Jonas Geiping, Yuxin Wen, Manli Shu, Khalid Saifullah, Kezhi Kong, Kasun Fernando, Aniruddha Saha, Micah Goldblum, and Tom Goldstein. 2023{\natexlab{b}}.
\newblock On the reliability of watermarks for large language models.
\newblock \emph{arXiv preprint arXiv:2306.04634}.

\bibitem[{Koo et~al.(2023)Koo, Lee, Raheja, Park, Kim, and Kang}]{koo2023benchmarking}
Ryan Koo, Minhwa Lee, Vipul Raheja, Jong~Inn Park, Zae~Myung Kim, and Dongyeop Kang. 2023.
\newblock Benchmarking cognitive biases in large language models as evaluators.
\newblock \emph{arXiv preprint arXiv:2309.17012}.

\bibitem[{Krishna et~al.(2023)Krishna, Song, Karpinska, Wieting, and Iyyer}]{krishna2023paraphrasing}
Kalpesh Krishna, Yixiao Song, Marzena Karpinska, John Wieting, and Mohit Iyyer. 2023.
\newblock Paraphrasing evades detectors of ai-generated text, but retrieval is an effective defense.
\newblock \emph{Advances in Neural Information Processing Systems}, 36:27469--27500.

\bibitem[{Kuditipudi et~al.(2023)Kuditipudi, Thickstun, Hashimoto, and Liang}]{kuditipudi2023robust}
Rohith Kuditipudi, John Thickstun, Tatsunori Hashimoto, and Percy Liang. 2023.
\newblock Robust distortion-free watermarks for language models.
\newblock \emph{arXiv preprint arXiv:2307.15593}.

\bibitem[{Kurita et~al.(2020)Kurita, Michel, and Neubig}]{kurita2020weight}
Keita Kurita, Paul Michel, and Graham Neubig. 2020.
\newblock Weight poisoning attacks on pre-trained models.
\newblock \emph{arXiv preprint arXiv:2004.06660}.

\bibitem[{Lee et~al.(2023)Lee, Hong, Ahn, Hong, Lee, Yun, Shin, and Kim}]{lee2023wrote}
Taehyun Lee, Seokhee Hong, Jaewoo Ahn, Ilgee Hong, Hwaran Lee, Sangdoo Yun, Jamin Shin, and Gunhee Kim. 2023.
\newblock Who wrote this code? watermarking for code generation.
\newblock \emph{arXiv preprint arXiv:2305.15060}.

\bibitem[{Liu et~al.(2024)Liu, Pan, Hu, Meng, and Wen}]{liu2024a}
Aiwei Liu, Leyi Pan, Xuming Hu, Shiao Meng, and Lijie Wen. 2024.
\newblock \href {https://openreview.net/forum?id=6p8lpe4MNf} {A semantic invariant robust watermark for large language models}.
\newblock In \emph{The Twelfth International Conference on Learning Representations}.

\bibitem[{Lou et~al.(2024)Lou, Liang, Xue, Zhang, Xie, and Zheng}]{lou2024cr}
Qian Lou, Xin Liang, Jiaqi Xue, Yancheng Zhang, Rui Xie, and Mengxin Zheng. 2024.
\newblock Cr-utp: Certified robustness against universal text perturbations on large language models.
\newblock In \emph{Findings of the Association for Computational Linguistics: ACL 2024}, pages 9863--9875.

\bibitem[{Lu et~al.(2024)Lu, Liu, Yu, Li, and King}]{lu2024entropy}
Yijian Lu, Aiwei Liu, Dianzhi Yu, Jingjing Li, and Irwin King. 2024.
\newblock An entropy-based text watermarking detection method.
\newblock \emph{arXiv preprint arXiv:2403.13485}.

\bibitem[{Mitchell et~al.(2023)Mitchell, Lee, Khazatsky, Manning, and Finn}]{mitchell_detectgpt_2023}
Eric Mitchell, Yoonho Lee, Alexander Khazatsky, Christopher~D. Manning, and Chelsea Finn. 2023.
\newblock \href {https://doi.org/10.48550/arXiv.2301.11305} {{{DetectGPT}}: {{Zero-Shot Machine-Generated Text Detection}} using {{Probability Curvature}}}.

\bibitem[{Pang et~al.(2024)Pang, Hu, Zheng, and Smith}]{pang2024no}
Qi~Pang, Shengyuan Hu, Wenting Zheng, and Virginia Smith. 2024.
\newblock No free lunch in llm watermarking: Trade-offs in watermarking design choices.
\newblock In \emph{The Thirty-eighth Annual Conference on Neural Information Processing Systems}.

\bibitem[{Papandreou and Yuille(2011)}]{papandreou2011perturb}
George Papandreou and Alan~L Yuille. 2011.
\newblock Perturb-and-map random fields: Using discrete optimization to learn and sample from energy models.
\newblock In \emph{2011 international conference on computer vision}, pages 193--200. IEEE.

\bibitem[{Pezeshkpour and Hruschka(2023)}]{pezeshkpour2023large}
Pouya Pezeshkpour and Estevam Hruschka. 2023.
\newblock Large language models sensitivity to the order of options in multiple-choice questions.
\newblock \emph{arXiv preprint arXiv:2308.11483}.

\bibitem[{Qi et~al.(2021)Qi, Li, Chen, Zhang, Liu, Wang, and Sun}]{qi2021hidden}
Fanchao Qi, Mukai Li, Yangyi Chen, Zhengyan Zhang, Zhiyuan Liu, Yasheng Wang, and Maosong Sun. 2021.
\newblock Hidden killer: Invisible textual backdoor attacks with syntactic trigger.
\newblock \emph{arXiv preprint arXiv:2105.12400}.

\bibitem[{Raffel et~al.(2020)Raffel, Shazeer, Roberts, Lee, Narang, Matena, Zhou, Li, and Liu}]{JMLR:v21:20-074}
Colin Raffel, Noam Shazeer, Adam Roberts, Katherine Lee, Sharan Narang, Michael Matena, Yanqi Zhou, Wei Li, and Peter~J. Liu. 2020.
\newblock \href {http://jmlr.org/papers/v21/20-074.html} {Exploring the limits of transfer learning with a unified text-to-text transformer}.
\newblock \emph{Journal of Machine Learning Research}, 21(140):1--67.

\bibitem[{{Sailor2 Team}(2024)}]{sailor2report}
{Sailor2 Team}. 2024.
\newblock Sailor2: Sailing in south-east asia with inclusive multilingual llm.

\bibitem[{Tian(2023)}]{tian_gptzero_2023}
Edward Tian. 2023.
\newblock \href {https://gptzero.substack.com/p/gptzero-update-v1} {Gptzero update v1}.

\bibitem[{Tiedemann et~al.(2022)Tiedemann, Aulamo, Bakshandaeva, Boggia, Gr{\"o}nroos, Nieminen, Raganato, Scherrer, Vazquez, and Virpioja}]{tiedemann2022democratizing}
J{\"o}rg Tiedemann, Mikko Aulamo, Daria Bakshandaeva, Michele Boggia, Stig-Arne Gr{\"o}nroos, Tommi Nieminen, Alessandro Raganato, Yves Scherrer, Raul Vazquez, and Sami Virpioja. 2022.
\newblock Democratizing machine translation with opus-mt.
\newblock \emph{arXiv preprint arXiv:2212.01936}.

\bibitem[{Tiedemann and Thottingal(2020)}]{tiedemann2020opus}
J{\"o}rg Tiedemann and Santhosh Thottingal. 2020.
\newblock {OPUS-MT} â {B}uilding open translation services for the {W}orld.
\newblock In \emph{Proceedings of the 22nd Annual Conference of the European Association for Machine Translation}.

\bibitem[{Xu et~al.(2024)Xu, Wang, Ma, Koh, Xiao, and Chen}]{xu2024instructional}
Jiashu Xu, Fei Wang, Mingyu~Derek Ma, Pang~Wei Koh, Chaowei Xiao, and Muhao Chen. 2024.
\newblock Instructional fingerprinting of large language models.
\newblock \emph{arXiv preprint arXiv:2401.12255}.

\bibitem[{Xue and Lou(2022)}]{xue2022estas}
Jiaqi Xue and Qian Lou. 2022.
\newblock Estas: Effective and stable trojan attacks in self-supervised encoders with one target unlabelled sample.
\newblock \emph{arXiv preprint arXiv:2211.10908}.

\bibitem[{Xue et~al.(2023{\natexlab{a}})Xue, Zheng, Hua, Shen, Liu, B{\"o}l{\"o}ni, and Lou}]{xue2023trojllm}
Jiaqi Xue, Mengxin Zheng, Ting Hua, Yilin Shen, Yepeng Liu, Ladislau B{\"o}l{\"o}ni, and Qian Lou. 2023{\natexlab{a}}.
\newblock Trojllm: A black-box trojan prompt attack on large language models.
\newblock \emph{Advances in Neural Information Processing Systems}, 36:65665--65677.

\bibitem[{Xue et~al.(2023{\natexlab{b}})Xue, Zheng, Sheng, Yang, Lou, and Jiang}]{xue2023trojfair}
Jiaqi Xue, Mengxin Zheng, Yi~Sheng, Lei Yang, Qian Lou, and Lei Jiang. 2023{\natexlab{b}}.
\newblock Trojfair: Trojan fairness attacks.
\newblock In \emph{Proceedings of the 1st ACM Workshop on Large AI Systems and Models with Privacy and Safety Analysis}, pages 47--56.

\bibitem[{Ye et~al.(2024)Ye, Wang, Huang, Chen, Zhang, Moniz, Gao, Geyer, Huang, Chen et~al.}]{ye2024justice}
Jiayi Ye, Yanbo Wang, Yue Huang, Dongping Chen, Qihui Zhang, Nuno Moniz, Tian Gao, Werner Geyer, Chao Huang, Pin-Yu Chen, et~al. 2024.
\newblock Justice or prejudice? quantifying biases in llm-as-a-judge.
\newblock \emph{arXiv preprint arXiv:2410.02736}.

\bibitem[{Zhao et~al.(2023)Zhao, Ananth, Li, and Wang}]{zhao2023provable}
Xuandong Zhao, Prabhanjan Ananth, Lei Li, and Yu-Xiang Wang. 2023.
\newblock Provable robust watermarking for ai-generated text.
\newblock \emph{arXiv preprint arXiv:2306.17439}.

\bibitem[{Zheng et~al.()Zheng, Zhou, Meng, Zhou, and Huang}]{zheng2309large}
Chujie Zheng, Hao Zhou, Fandong Meng, Jie Zhou, and Minlie Huang.
\newblock Large language models are not robust multiple choice selectors, 2024.
\newblock \emph{URL https://arxiv. org/abs/2309.03882}.

\bibitem[{Zheng et~al.(2023)Zheng, Lou, and Jiang}]{zheng2023trojvit}
Mengxin Zheng, Qian Lou, and Lei Jiang. 2023.
\newblock Trojvit: Trojan insertion in vision transformers.
\newblock In \emph{Proceedings of the IEEE/CVF Conference on Computer Vision and Pattern Recognition}, pages 4025--4034.

\bibitem[{Zheng et~al.(2024)Zheng, Xue, Chen, Wang, Lou, and Jiang}]{zheng2024trojfsp}
Mengxin Zheng, Jiaqi Xue, Xun Chen, Yanshan Wang, Qian Lou, and Lei Jiang. 2024.
\newblock Trojfsp: Trojan insertion in few-shot prompt tuning.
\newblock In \emph{Proceedings of the 2024 Conference of the North American Chapter of the Association for Computational Linguistics: Human Language Technologies (Volume 1: Long Papers)}, pages 1141--1151.

\bibitem[{Zhu et~al.(2018)Zhu, Lu, Zheng, Guo, Zhang, Wang, and Yu}]{zhu2018texygen}
Yaoming Zhu, Sidi Lu, Lei Zheng, Jiaxian Guo, Weinan Zhang, Jun Wang, and Yong Yu. 2018.
\newblock Texygen: A benchmarking platform for text generation models.
\newblock \emph{SIGIR}.

\end{thebibliography}
\newpage
\appendix
\section{More Quality Results}\label{sec:appendix-quality}
\subsection{Perplexity (PPL)}
Perplexity is a metric that, although known for its limitations, still provides a useful statistical or probabilistic background for assessing text quality. It measures how well a probability distribution predicts a sample. In the context of language models, it provides insights into how \emph{surprised} the model is by the actual sequence of words compared to its predicted probabilities. A lower perplexity score indicates that the model found the text more predictable and thus, in a broad sense, of higher quality. In our experiments, the PPL is calculated as follows:
\[
\text{Perplexity}(W) = \exp\left(\frac{1}{N} \sum_{i=1}^{N} -\log p(w_i)\right)
\]
where \( \exp \) represents the exponential function. \( N \) is the total number of words in the text. \( p(w_i) \) is the probability assigned to the word \( w_i \) by the model. The negative log likelihood, \(-\log p(w_i)\), is referred to the Cross-Entropy loss for word \( w_i \).

\textbf{PPL Analysis}:
\begin{figure*}[th]
  \subfigure[KGW]{\includegraphics[width=0.98\columnwidth]{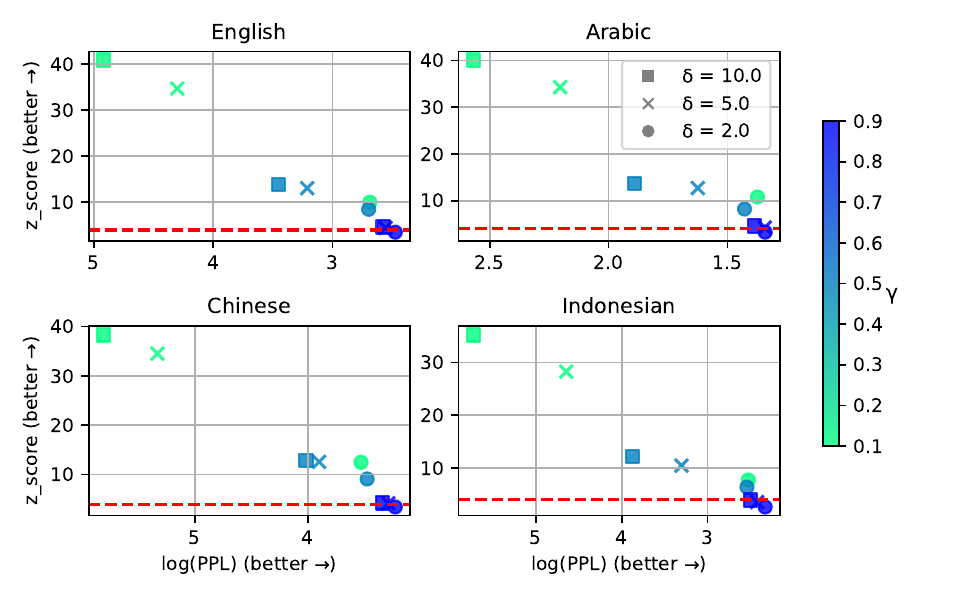}\label{fig:z-vs-ppl-kgw}} \hfill
  \subfigure[Unigram]{\includegraphics[width=0.98\columnwidth]{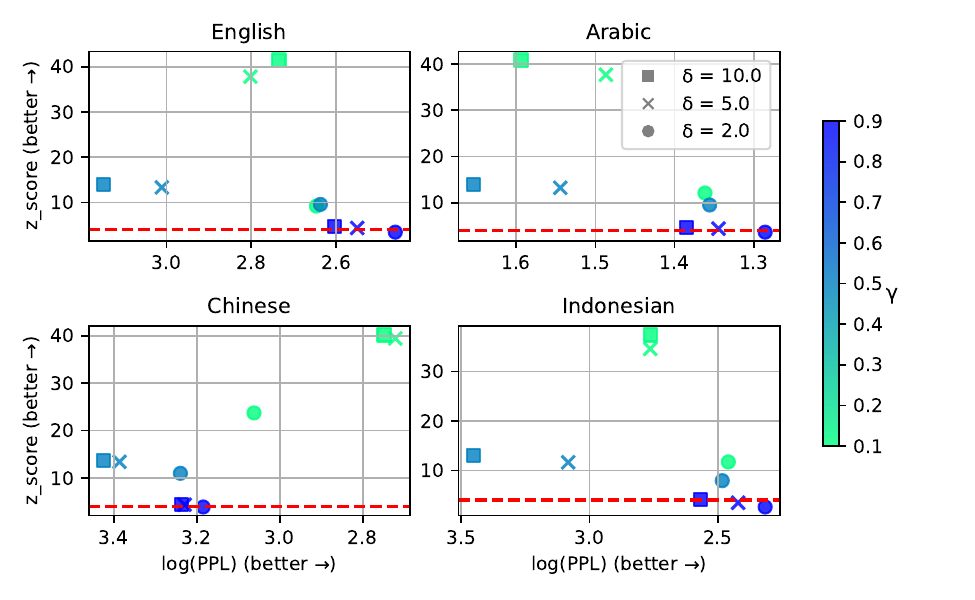}\label{fig:z-vs-ppl-unigram}}
  \vspace{-3mm}
  \caption {Detection scores as a function of PPL for KGW and Unigram. For KGW, we notice that as $\delta$ grows higher, the quality of text decreases for all languages. Larger $\gamma$ values with smaller $\delta$ values greately affected the watermark strength in which it is attenuated. Unigram presents interesting graphs. When $\gamma$ is large, we see simiar trend as in KGW. However, smaller $\gamma$ values behave differently for different languages. The PPL and Z-scores are calculated on 500 generations.}
  \label{fig:z-scores-vs-ppls-kgw-unigram}
  \vspace{-3mm}
\end{figure*}

\begin{figure*}[th]
  \subfigure[XSIR]{\includegraphics[width=0.98\columnwidth]{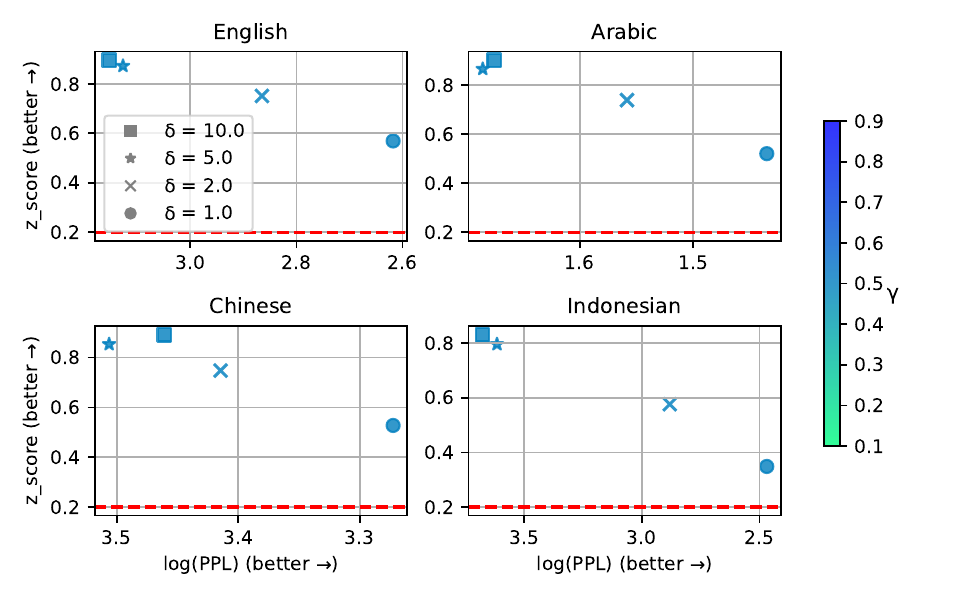}\label{fig:z-vs-ppl-xsir}} \hfill
  \subfigure[EXP]{\includegraphics[width=0.98\columnwidth]{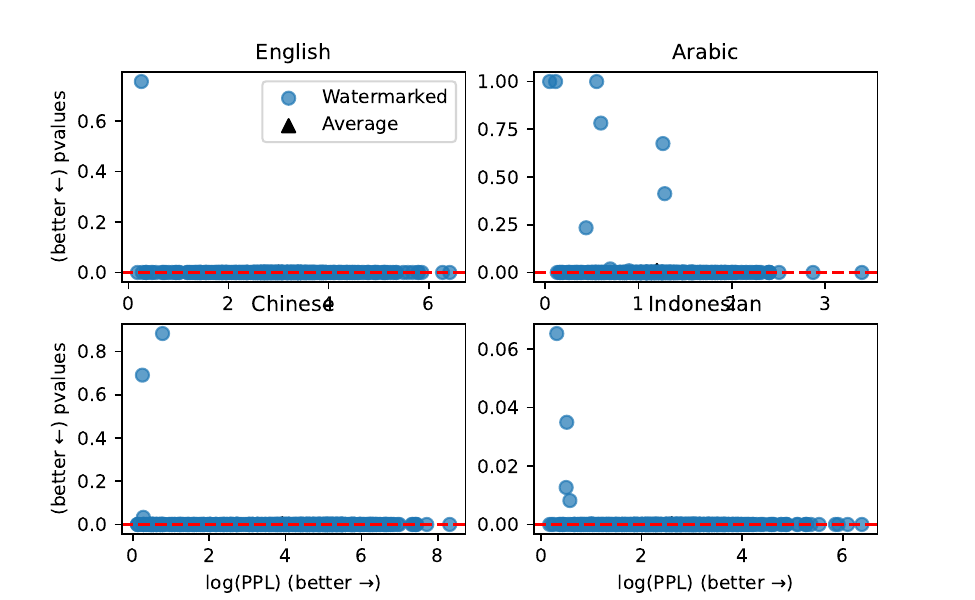}\label{fig:z-vs-ppl-exp}}
  \vspace{-3mm}
  \caption {Detection scores as a function of PPL for XSIR and EXP. XSIR employs $\gamma=0.5$. The smaller the $\delta$ the better the qulaity of text. EXP $p$-values show insensitivity to PPL scores with a few examples falling above the threshold or indicating False Negatives.}
  \label{fig:z-scores-vs-ppls-others}
  \vspace{-3mm}
\end{figure*}
Figures ~\ref{fig:z-scores-vs-ppls-kgw-unigram} and ~\ref{fig:z-scores-vs-ppls-others} show the strength of detection of the watermark as a function of PPL. The results in figure ~\ref{fig:z-vs-ppl-kgw} confirm the findings in ~\cite{kirchenbauer2023watermark} about the slight effect of larger $\delta$ values to the quality of the watermarked text for all languages. We see similar trends for Unigram in figure ~\ref{fig:z-vs-ppl-unigram} only when $\gamma \geq 0.5$. For smaller values of $\gamma$, the quality of text seems to slightly improve even for larger $\delta$ values. We believe that this effect is a result of low-entropy completions from the red list being repeated over and over given smaller potion of the vocabulary, which increases the z-scores and precludes PPL from catching this effect.

For the other watermark methods XSIR and EXP, we show their results in figure ~\ref{fig:z-scores-vs-ppls-others}. For EXP, we experiment with the whole data points to clearly identify the trends. In figure ~\ref{fig:z-vs-ppl-xsir}, XSIR shows similar trends to KGW in terms of the effect of larger values of $\delta$ on text quality. For EXP as shown in figure ~\ref{fig:z-vs-ppl-exp}, the quality of text has a consistent relationship with the lower $p$-values across all languages. In other words, the $p$-values remain virtually insensitive to variations in the quality of the text.

\subsection{Complete Self-BLEU Results}
Table ~\ref{tab:self-bleu-more}, we show complete results of self-BLEU and Adjusted Diversity (AD) using different hyper-parameter settings. From the results, it's clear that when the $\delta$ is large, the text diversity unrealistically increases for KGW, Unigram and XSIR. This is due to the tight relationship between $\gamma$ and $\delta$ in which lower $\gamma$ values with higher $\delta$ values causes an adverse effect on the text quality.

\subsection{Sensitivity Analysis for $w$ choice in AD metric}
The choice of $w$ in our paper is based on empirical results. In Table~\ref{tab:sb-ad-results} we show more resuts with varying the $w$ parameters to be in $[0.1, 0.3, 0.7]$.
After we generate the outputs for a specific language, and after we noticed unexplainable numbers for some results (for example very low Self-Bleu scores), we revise the outputs to check if something is off with the data. Self-blue (SB) results are sometimes $ < 0.05 $ which is good in terms of diversity, yet the quality of the output is not as good due to mixture of characters from different languages or gibberish chars, hence lower values of SB scores. However, when the text is generated correctly, we want to catch any repetition in the text due to watermarking. The result of the Judge could still help here since we only utilize its explanation about the coherency of the text, and not its final verdict as to which text is better to avoid judger choice biases. For this reason, we preferred higher weight when we used the coherency scores. Additionally, larger $w$ weights will attribute more weight to SB which is mere n-gram test. Therefore, any non-repeated but incorrect generations will not be explained in the SB scores.

\begin{table*}[th!]
    \centering
    \small
    \caption{Complete SelfBleu results with the Adjusted Diversity (AD) metric with $w=0.3$ for all watermark methods. For KGW, Unigram, and XSIR, we fix $\gamma=0.5$ and vary $\delta=[2.0, 5.0, 10.0]$}
    \begin{tabular}{cccccccc}
        \toprule
        Method & Language & $\gamma$ & $\delta$ & \multicolumn{2}{c}{self-bleu ($\downarrow$ more diverse)} & \multicolumn{2}{c}{AD ($\downarrow$ more diverse)}\\
        \cmidrule(lr){5-6}
        \cmidrule(lr){7-8}
        & & & & watermarked & unwatermarked & watermarked & unwatermarked\\
        \midrule
        \multirow{4}{*}{KGW} & \multirow{3}{*}{English}
        & \multirow{3}{*}{0.5} & 2.0 & 0.16 & 0.16 & 0.38 & 0.34\\
        & & & 5.0 & 0.15 & 0.17 & 0.46 & 0.32\\
        & & & 10.0 & 0.15 & 0.17 & \textbf{0.50} & 0.31\\
        \cmidrule(lr){2-8}
        & \multirow{3}{*}{Arabic}
        & \multirow{3}{*}{0.5} & 2.0 & 0.11 & 0.12 & 0.36 & 0.35\\
        & & & 5.0 & 0.12 & 0.12 & 0.41 & 0.35\\
        & & & 10.0 & 0.12 & 0.13 & \textbf{0.45} & 0.33\\
        \cmidrule(lr){2-8}
        & \multirow{3}{*}{Chinese}
        & \multirow{3}{*}{0.5} & 2.0 & 0.04 & 0.04 & 0.44 & 0.40\\
        & & & 5.0 & 0.04 & 0.04 & 0.46 & 0.41\\
        & & & 10.0 & 0.04 & 0.04 & \textbf{0.48} & 0.40\\
        \cmidrule(lr){2-8}
        & \multirow{3}{*}{Indonesian}
        & \multirow{3}{*}{0.5} & 2.0 & 0.10 & 0.10 & 0.43 & 0.38\\
        & & & 5.0 & 0.08 & 0.10 & 0.53 & 0.35\\
        & & & 10.0 & 0.08 & 0.11 & \textbf{0.60} & 0.33\\
        \midrule
        \multirow{4}{*}{Unigram} & \multirow{3}{*}{English}
        & \multirow{3}{*}{0.5} & 2.0 & 0.23 & 0.17 & 0.40 & 0.35\\
        & & & 5.0 & 0.26 & 0.17 & 0.54 & 0.32\\
        & & & 10.0 & 0.27 & 0.17 & \textbf{0.59} & 0.31\\
        \cmidrule(lr){2-8}
        & \multirow{3}{*}{Arabic}
        & \multirow{3}{*}{0.5} & 2.0 & 0.16 & 0.12 & 0.41 & 0.35\\
        & & & 5.0 & 0.19 & 0.12 & 0.48 & 0.32\\
        & & & 10.0 & 0.21 & 0.12 & \textbf{0.52} & 0.34\\
        \cmidrule(lr){2-8}
        & \multirow{3}{*}{Chinese}
        & \multirow{3}{*}{0.5} & 2.0 & 0.02 & 0.04 & 0.47 & 0.40\\
        & & & 5.0 & 0.02 & 0.04 & 0.52 & 0.39\\
        & & & 10.0 & 0.03 & 0.03 & \textbf{0.54} & 0.39\\
        \cmidrule(lr){2-8}
        & \multirow{3}{*}{Indonesian}
        & \multirow{3}{*}{0.5} & 2.0  & 0.12 & 0.10 & 0.46 & 0.37\\
        & & & 5.0 & 0.13 & 0.10 & 0.57 & 0.34\\
        & & & 10.0 & 0.13 & 0.11 & \textbf{0.63} & 0.33\\
        \midrule
        \multirow{4}{*}{XSIR} & \multirow{3}{*}{English}
        & \multirow{3}{*}{0.5} & 2.0 & 0.20 & 0.17  & 0.49 & 0.32\\
        & & & 5.0 & 0.22 & 0.17 & 0.59 & 0.30\\
        & & & 10.0 & 0.22 & 0.17 & \textbf{0.63} & 0.29\\
        \cmidrule(lr){2-8}
        & \multirow{3}{*}{Arabic}
        & \multirow{3}{*}{0.5} & 2.0 & 0.14 & 0.12 & 0.45 & 0.33\\
        & & & 5.0 & 0.19 & 0.13 & 0.55 & 0.30\\
        & & & 10.0 & 0.19 & 0.13 & \textbf{0.57} & 0.31\\
        \cmidrule(lr){2-8}
        & \multirow{3}{*}{Chinese}
        & \multirow{3}{*}{0.5} & 2.0 & 0.03 & 0.04 & 0.55 & 0.38\\
        & & & 5.0 & 0.04 & 0.04 & 0.61 & 0.36\\
        & & & 10.0 & 0.03 & 0.04 & \textbf{0.62} & 0.36\\
        \cmidrule(lr){2-8}
        & \multirow{3}{*}{Indonesian}
        & \multirow{3}{*}{0.5} & 2.0 & 0.12 & 0.10 & 0.53 & 0.37\\
        & & & 5.0 & 0.10 & 0.10 & 0.66 & 0.32\\
        & & & 10.0 & 0.10 & 0.10 & \textbf{0.68} & 0.31\\
        \midrule
        \multirow{4}{*}{EXP} & English
        & \multirow{4}{*}{-} & \multirow{4}{*}{-} & 0.19 & 0.17 & 0.57 & 0.29 \\
        & Arabic
        &  &  & 0.20 & 0.12 & 0.55 & 0.31\\
        & Chinese
        &  &  & 0.13 & 0.04 & \textbf{0.61} & 0.38\\
        & Indonesian
        &  &  & 0.21 & 0.10 & 0.57 & 0.36\\
        \bottomrule
    \end{tabular}
    \label{tab:self-bleu-more}
\end{table*}

\begin{table*}[th!]
    \centering
    \small
    \caption{Self-BLEU / Adjusted Diversity for All Methods, Languages, and Weights in $w=[0.1, 0.3, 0.7]$ for CohereAI model. The $w$ weight is used in equation~\ref{eq:AD-metric} where smaller values indicate larger weighing from judger coherency scores. Smaller values like 0.1 and 0.3 seems to reflect closer AD scores for the watermarked text in comparison to a large one such as 0.7.}
    \label{tab:sb-ad-results}
    \begin{tabular}{llccc}
        \toprule
        Method & Lang & $w$ & Watermarked SB / AD & Unwatermarked SB / AD \\
        \midrule
        \multirow{4}{*}{KGW} & En & 0.1 & 0.16 / 0.44 & 0.16 / 0.40 \\
        &    & 0.3 & 0.16 / 0.38 & 0.16 / 0.34 \\
        &    & 0.7 & 0.16 / 0.25 & 0.16 / 0.24 \\
        \cmidrule(lr){2-5}
        & Ar & 0.1 & 0.11 / 0.43 & 0.12 / 0.42 \\
        &    & 0.3 & 0.11 / 0.36 & 0.12 / 0.35 \\
        &    & 0.7 & 0.11 / 0.22 & 0.12 / 0.22 \\
        \cmidrule(lr){2-5}
        & Zh & 0.1 & 0.04 / 0.55 & 0.04 / 0.51 \\
        &    & 0.3 & 0.04 / 0.44 & 0.04 / 0.40 \\
        &    & 0.7 & 0.04 / 0.21 & 0.04 / 0.19 \\
        \cmidrule(lr){2-5}
        & Id & 0.1 & 0.10 / 0.52 & 0.10 / 0.46 \\
        &    & 0.3 & 0.10 / 0.43 & 0.10 / 0.38 \\
        &    & 0.7 & 0.10 / 0.24 & 0.10 / 0.22 \\
        \midrule
        \multirow{4}{*}{Unigram} & En & 0.1 & 0.23 / 0.45 & 0.17 / 0.40 \\
        &    & 0.3 & 0.23 / 0.40 & 0.17 / 0.35 \\
        &    & 0.7 & 0.23 / 0.30 & 0.17 / 0.24 \\
        \cmidrule(lr){2-5}
        & Ar & 0.1 & 0.16 / 0.47 & 0.12 / 0.42 \\
        &    & 0.3 & 0.16 / 0.41 & 0.12 / 0.35 \\
        &    & 0.7 & 0.16 / 0.27 & 0.12 / 0.22 \\
        \cmidrule(lr){2-5}
        & Zh & 0.1 & 0.02 / 0.60 & 0.04 / 0.51 \\
        &    & 0.3 & 0.02 / 0.47 & 0.04 / 0.40 \\
        &    & 0.7 & 0.02 / 0.21 & 0.04 / 0.20 \\
        \cmidrule(lr){2-5}
        & Id & 0.1 & 0.12 / 0.55 & 0.10 / 0.45 \\
        &    & 0.3 & 0.12 / 0.46 & 0.10 / 0.37 \\
        &    & 0.7 & 0.12 / 0.27 & 0.10 / 0.22 \\
        \midrule
        \multirow{4}{*}{XSIR} & En & 0.1 & 0.20 / 0.57 & 0.17 / 0.37 \\
        &    & 0.3 & 0.20 / 0.49 & 0.17 / 0.32 \\
        &    & 0.7 & 0.20 / 0.32 & 0.17 / 0.24 \\
        \cmidrule(lr){2-5}
        & Ar & 0.1 & 0.14 / 0.54 & 0.12 / 0.38 \\
        &    & 0.3 & 0.14 / 0.45 & 0.12 / 0.33 \\
        &    & 0.7 & 0.14 / 0.27 & 0.12 / 0.21 \\
        \cmidrule(lr){2-5}
        & Zh & 0.1 & 0.03 / 0.70 & 0.04 / 0.48 \\
        &    & 0.3 & 0.03 / 0.55 & 0.04 / 0.38 \\
        &    & 0.7 & 0.03 / 0.25 & 0.04 / 0.19 \\
        \cmidrule(lr){2-5}
        & Id & 0.1 & 0.12 / 0.65 & 0.11 / 0.44 \\
        &    & 0.3 & 0.12 / 0.53 & 0.11 / 0.37 \\
        &    & 0.7 & 0.12 / 0.29 & 0.11 / 0.22 \\
        \midrule
        \multirow{4}{*}{EXP} & En & 0.1 & 0.19 / 0.67 & 0.17 / 0.33 \\
        &    & 0.3 & 0.19 / 0.57 & 0.17 / 0.29 \\
        &    & 0.7 & 0.19 / 0.35 & 0.17 / 0.23 \\
        \cmidrule(lr){2-5}
        & Ar & 0.1 & 0.20 / 0.65 & 0.12 / 0.37 \\
        &    & 0.3 & 0.20 / 0.55 & 0.12 / 0.31 \\
        &    & 0.7 & 0.20 / 0.35 & 0.12 / 0.21 \\
        \cmidrule(lr){2-5}
        & Zh & 0.1 & 0.13 / 0.74 & 0.04 / 0.48 \\
        &    & 0.3 & 0.13 / 0.61 & 0.04 / 0.38 \\
        &    & 0.7 & 0.13 / 0.34 & 0.04 / 0.19 \\
        \cmidrule(lr){2-5}
        & Id & 0.1 & 0.21 / 0.67 & 0.10 / 0.43 \\
        &    & 0.3 & 0.21 / 0.57 & 0.10 / 0.36 \\
        &    & 0.7 & 0.21 / 0.36 & 0.10 / 0.21 \\
        \bottomrule
\end{tabular}
\end{table*}
\subsection{Complete Soft-Win Results}
In Table~\ref{tab:soft-win-rates-more}, we show more results across different hyper-parameter settings for KGW, Unigram and XSIR. The soft-win rates drastically decreases with lower $\gamma$ values and higher $\delta$ values. This is explainable since larger $\delta$ values increases the magnitude of watermark signal in which the next-token generation is greatly affected.

\begin{table*}[th!]
    \centering
    \small
    \caption{Soft Win Rates for Different Methods by Language and Hyper-parameters. Columns are added for averages across languages and methods.}
    \begin{tabular}{ccccccc}
        \toprule
        Method & Language & $\gamma$ & $\delta$ & Soft Win Rate & Language Avg. & Method Avg. \\
        \midrule
        \multirow{9}{*}{KGW} & \multirow{3}{*}{English}
        & \multirow{3}{*}{0.5} & 2.0 & 0.47 & \multirow{3}{*}{0.314} & \multirow{12}{*}{0.355} \\
        & & & 5.0 & 0.264 & & \\
        & & & 10.0 & 0.21 & & \\
        \cmidrule(lr){2-6}
        & \multirow{3}{*}{Arabic}
        & \multirow{3}{*}{0.5} & 2.0 & 0.56 & \multirow{3}{*}{0.417} & \\
        & & & 5.0 & 0.418 & & \\
        & & & 10.0 & 0.314 & & \\
        \cmidrule(lr){2-6}
        & \multirow{3}{*}{Chinese}
        & \multirow{3}{*}{0.5} & 2.0 & 0.566 & \multirow{3}{*}{0.502} & \\
        & & & 5.0 & 0.514 & & \\
        & & & 10.0 & 0.426 & & \\
        \cmidrule(lr){2-6}
        & \multirow{3}{*}{Indonesian}
        & \multirow{3}{*}{0.5} & 2.0 & 0.468 & \multirow{3}{*}{0.267} & \\
        & & & 5.0 & 0.224 & & \\
        & & & 10.0 & 0.108 & & \\
        \midrule
        \multirow{9}{*}{Unigram} & \multirow{3}{*}{English}
        & \multirow{3}{*}{0.5} & 2.0 & 0.462 & \multirow{3}{*}{0.287} & \multirow{12}{*}{0.314} \\
        & & & 5.0 & 0.232 & & \\
        & & & 10.0 & 0.166 & & \\
        \cmidrule(lr){2-6}
        & \multirow{3}{*}{Arabic}
        & \multirow{3}{*}{0.5} & 2.0 & 0.494 & \multirow{3}{*}{0.352} & \\
        & & & 5.0 & 0.292 & & \\
        & & & 10.0 & 0.27 & & \\
        \cmidrule(lr){2-6}
        & \multirow{3}{*}{Chinese}
        & \multirow{3}{*}{0.5} & 2.0 & 0.51 & \multirow{3}{*}{0.436} & \\
        & & & 5.0 & 0.422 & & \\
        & & & 10.0 & 0.376 & & \\
        \cmidrule(lr){2-6}
        & \multirow{3}{*}{Indonesian}
        & \multirow{3}{*}{0.5} & 2.0 & 0.412 & \multirow{3}{*}{0.234} & \\
        & & & 5.0 & 0.178 & & \\
        & & & 10.0 & 0.112 & & \\
        \midrule
        \multirow{9}{*}{XSIR} & \multirow{3}{*}{English}
        & \multirow{3}{*}{0.5} & 2.0 & 0.342 & \multirow{3}{*}{0.256} & \multirow{12}{*}{0.218} \\
        & & & 5.0 & 0.2 & & \\
        & & & 10.0 & 0.226 & & \\
        \cmidrule(lr){2-6}
        & \multirow{3}{*}{Arabic}
        & \multirow{3}{*}{0.5} & 2.0 & 0.26 & \multirow{3}{*}{0.183} & \\
        & & & 5.0 & 0.136 & & \\
        & & & 10.0 & 0.1 & & \\
        \cmidrule(lr){2-6}
        & \multirow{3}{*}{Chinese}
        & \multirow{3}{*}{0.5} & 2.0 & 0.35 & \multirow{3}{*}{0.293} & \\
        & & & 5.0 & 0.262 & & \\
        & & & 10.0 & 0.268 & & \\
        \cmidrule(lr){2-6}
        & \multirow{3}{*}{Indonesian}
        & \multirow{3}{*}{0.5} & 2.0 & 0.276 & \multirow{3}{*}{0.15} & \\
        & & & 5.0 & 0.108 & & \\
        & & & 10.0 & 0.066 & & \\
        \midrule
        \multirow{4}{*}{EXP} & English
        & \multirow{4}{*}{-} & \multirow{4}{*}{-} & 0.142 & \multirow{4}{*}{-} & \multirow{4}{*}{0.24} \\
        & Arabic
        &  &  & 0.246 & & \\
        & Chinese
        &  &  & 0.31 & & \\
        & Indonesian
        &  &  & 0.262 & & \\
        \bottomrule
    \end{tabular}
    \label{tab:soft-win-rates-more}
\end{table*}

\subsection{More Quality Results for Turkish and Hindi Languages}
\begin{figure*}[h]
\centering
\includegraphics[width=\linewidth]{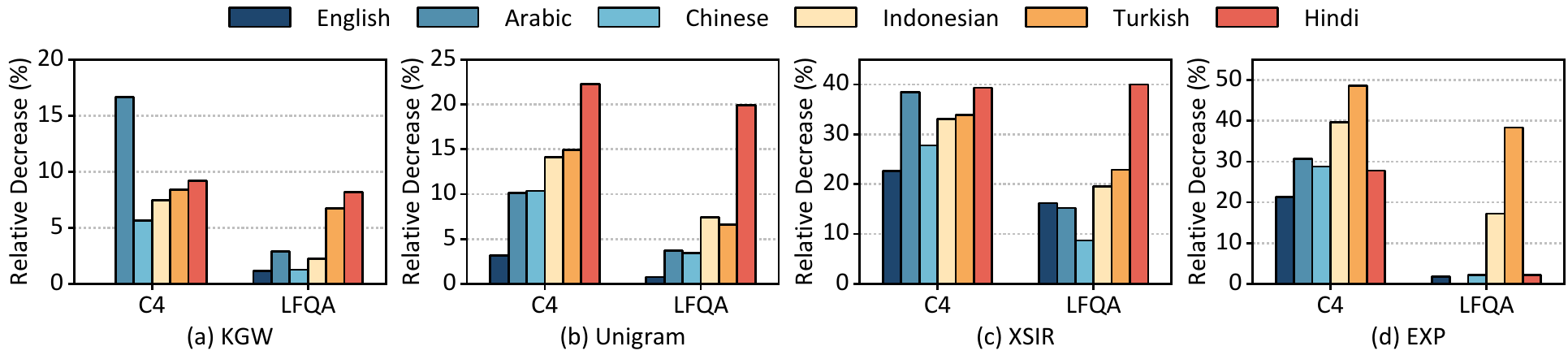}
\caption{GPT-judge coherency criterion results. We compute the average of watermarked and unwatermarked scores for $500$ generations. Here we only use Cohere model for more languages like Turkish and Hindi.}
\label{fig:gptjudge-results-more-langs}
\end{figure*}

In Figure~\ref{fig:gptjudge-results-more-langs}, we show more coherency difference results for low-resource languages such as Turkish and Hindi using the CohereAI model.

\section{More Detection Results}\label{sec:appendix-detection}
\begin{figure*}[th]
  \subfigure[KGW]{\includegraphics[width=0.5\columnwidth]{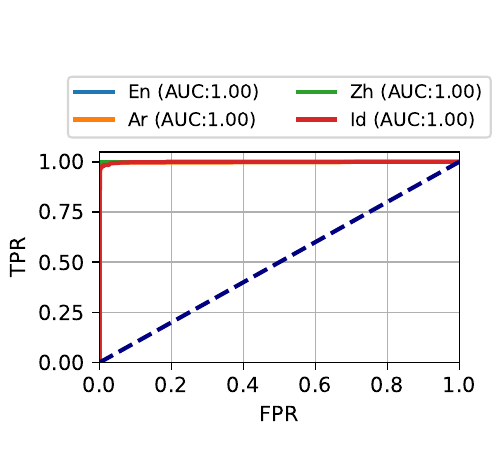}\label{fig:roc-curve-kgw-before}}
  \subfigure[Unigram]{\includegraphics[width=0.5\columnwidth]{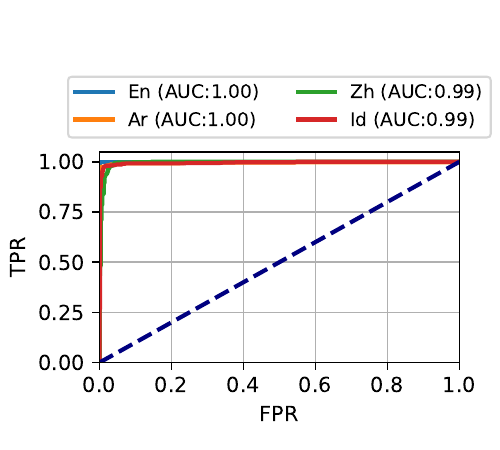}\label{fig:roc-curve-unigram-before}} \hfill
  \subfigure[XSIR]{\includegraphics[width=0.5\columnwidth]{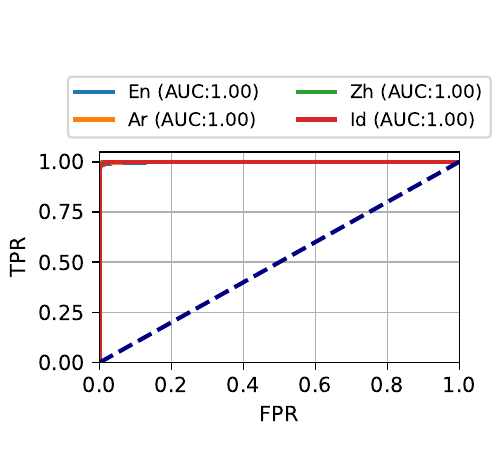}\label{fig:roc-curve-xsir-before}}
  \subfigure[EXP]{\includegraphics[width=0.5\columnwidth]{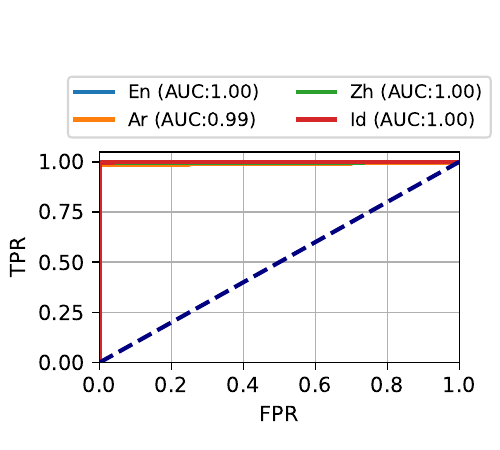}\label{fig:roc-curve-exp-before}}
  \vspace{-3mm}
  \caption {Watermark detection ROC curves with AUC before attacks. We fix $\gamma=0.5$ and $\delta=2.0$ for KGW, Unigram, and XSIR. The watermark threshold is calculated automatically by comparing unwatermarked and watermarked scores for $500$ generations.}
  \label{fig:roc-curves-all-methods-before}
  \vspace{-3mm}
\end{figure*}

\begin{figure*}[th!]
  \subfigure[KGW]{\includegraphics[width=0.98\columnwidth]{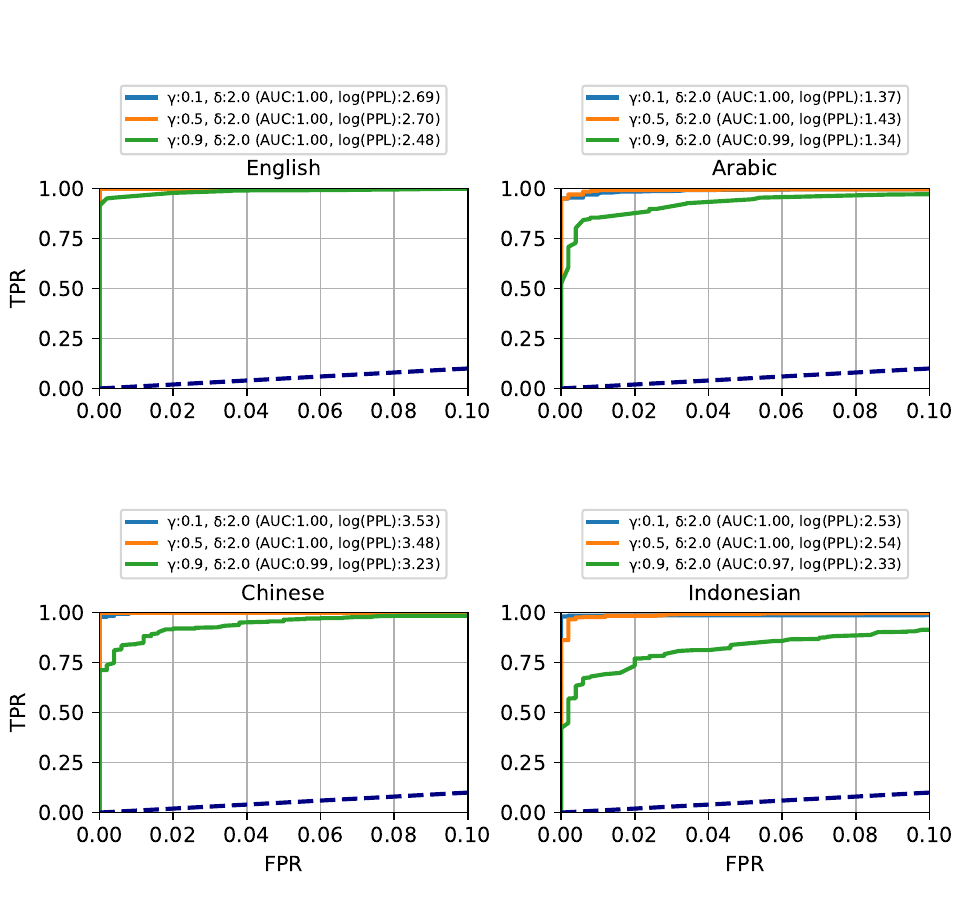}\label{fig:roc-curve-kgw-before-zoomed}} \hfill
  \subfigure[Unigram]{\includegraphics[width=0.98\columnwidth]{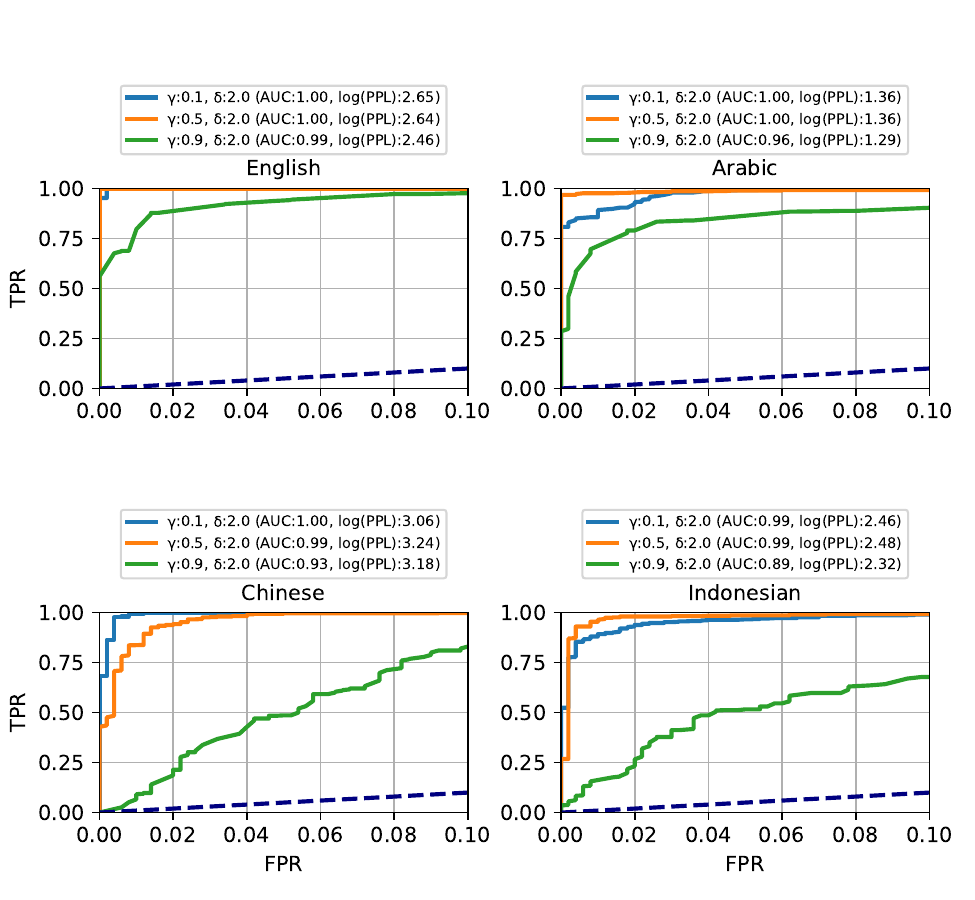}\label{fig:roc-curve-unigram-before-zoomed}}
  \vspace{-3mm}
  \caption {Here we fix $\delta$ at $2.0$ and vary $\gamma$ for KGW and Unigram with $\gamma=(0.1, 0.5, 0.9)$ and with a fixed $\delta=2.0$. The larger the $\gamma$ ratio, the worse the watermark detection. We also use smaller FPRs in the range $[0.1, 0.15]$}
  \label{fig:roc-curves-kgw-unigram-before-zoomed}
  \vspace{-3mm}
\end{figure*}

\begin{figure*}[th!]
  \subfigure[XSIR]{\includegraphics[width=0.98\columnwidth]{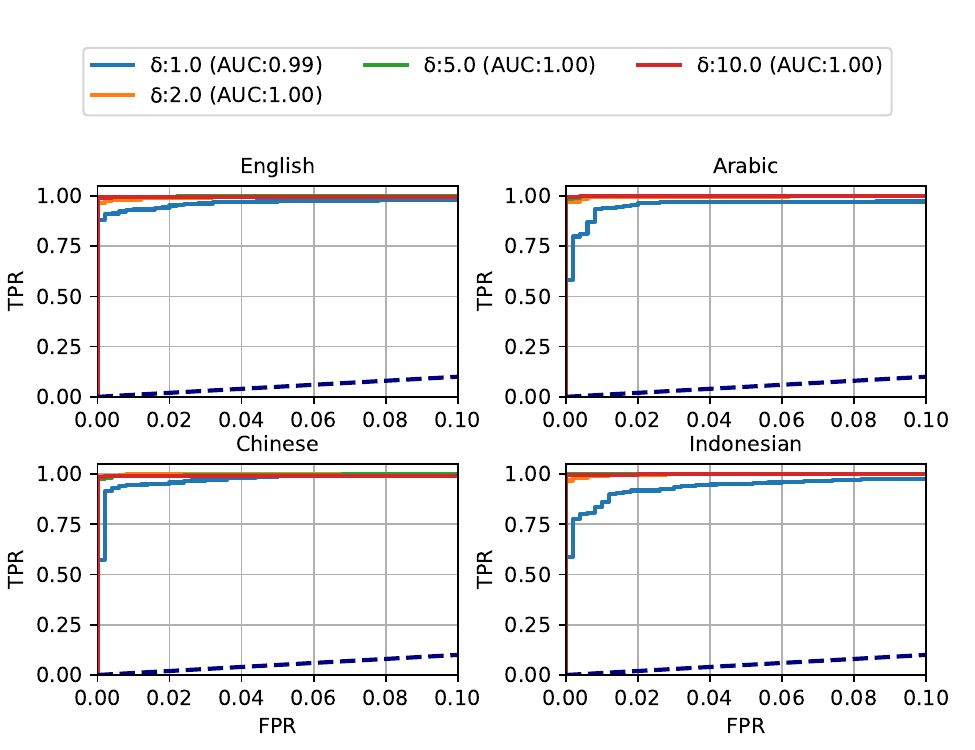}\label{fig:roc-curve-xsir-zoomed}} \hfill
  \subfigure[EXP]{\includegraphics[width=0.98\columnwidth]{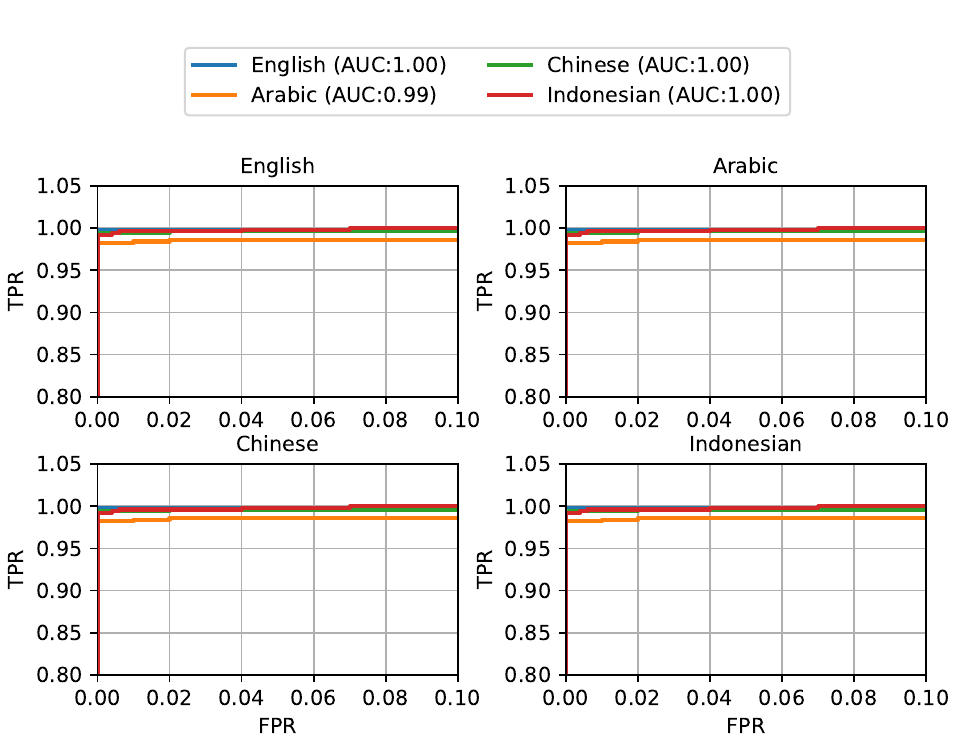}\label{fig:roc-curve-exp-zoomed}}
  \vspace{-3mm}
  \caption {Watermark detection for XSIR and EXP with lower values of FPRs, which are  in the range $[0.1, 0.15]$. For XSIR, lower values of $\delta$ results in lower TPRs at very low FPRs.}
  \label{fig:roc-curves-xsir-exp-before-zoomed}
  \vspace{-3mm}
\end{figure*}

\begin{figure*}[th]
  \subfigure[Translation->Paraphrase Attacks]{\includegraphics[width=0.98\columnwidth]{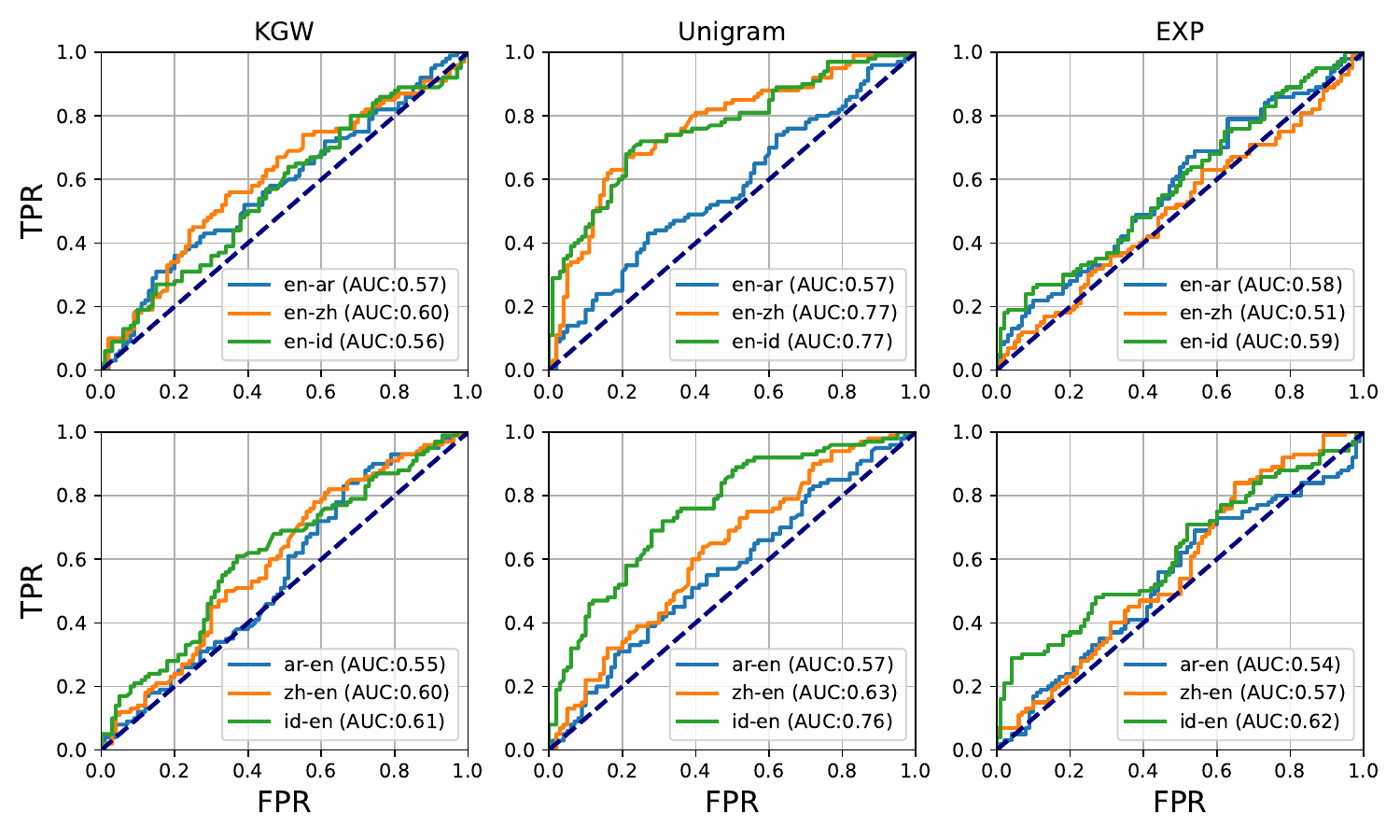}\label{fig:roc-curve-kgw-unigram-exp-translations->paraphrase}} \hfill
  \subfigure[Translation->Paraphrase->Translation Attacks]{\includegraphics[width=0.98\columnwidth]{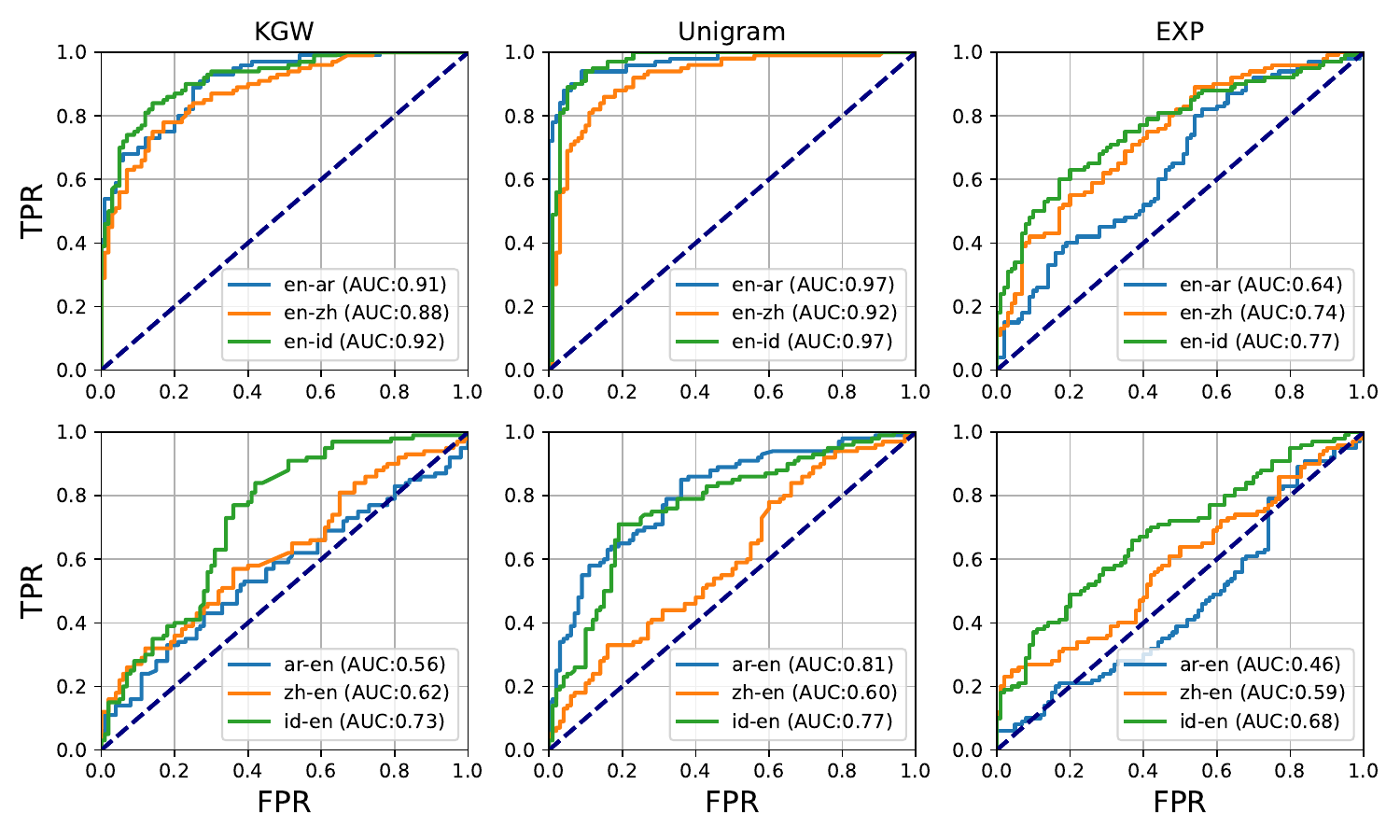}\label{fig:roc-curve-kgw-unigram-exp-all-attacks}}
  \vspace{-3mm}
  \caption {Watermark detection ROC curves with AUC after attacks for syntactical methods KGW, Unigram, and EXP. We fix $\gamma=0.5$ and $\delta=2.0$ for KGW and Unigram. For each attack, the watermark threshold is calculated automatically by comparing unwatermarked and attacked watermarked text score for $100$ generations. The bottom row in all figures represent XSIR original attack setting in which non-English language is used first.}
  \label{fig:roc-curves-syntactical-methods-after}
  \vspace{-3mm}
\end{figure*}
\subsection{Watermark Detection with Multiple $\gamma$ Ratios for KGW and Unigram}
KGW ~\citet{kirchenbauer2023watermark} as shown in ~\ref{fig:roc-curve-kgw-before} performs the best among all other watermark methods, while Unigram as shown in figure ~\ref{fig:roc-curve-unigram-before} performs the worst. XSIR and EXP as shown in figures ~\ref{fig:roc-curve-xsir-before} and ~\ref{fig:roc-curve-exp-before} respectively show higher TPRs compared to KGW and Unigram. In all watermark methods, English language shows stable AUC curve in all FPRs in all watermark methods. For KGW and Unigram, increasing the value of $\gamma$ and decreasing the value of $\delta$ has adverse effect on the TPR characteristic, especially for languages other than English (see subsequent sections for roc curves with zoomed-in rates). 

\citet{kirchenbauer2023watermark} has investigated the tight relationship of $\gamma$ and $\delta$ in watermark strength by creating a lower bound on $\gamma$ with the spike entropy in the picture. However, we believe more investigation is needed in cross-lingual manner. For example, beside the effect of the sampling method (greedy or multinomial), what could be done to incorporate the role of the tokenization of different languages in the $\gamma$ lower bound analysis? By doing a simple analysis of the results with $\gamma=0.9$ and $\delta=0.2$, we found that only an average of $2$ greenlist tokens are missing to raise the z-score from, say $3.52$ to $4$, with a z-score of threshold $4$. As shown in Figures \ref{fig:roc-curve-kgw-before-zoomed} and \ref{fig:roc-curve-unigram-before-zoomed}, the z-score threshold performs adequately for both methods in English but struggles with other languages. KGW experiences TPR drop in Arabic and Indonesian, while Unigram's TPR drops primarily in Chinese and Indonesian with low FPR rates. XSIR as shown in Figure \ref{fig:roc-curve-xsir-zoomed} has relatively minor effects for smaller values of $\delta$. This is because XSIR logically uses $50\%$ of the total vocabulary, which is high compared to a smaller $\delta$ value such as $2.0$. Finally, EXP, as shown in Figure ~\ref{fig:roc-curve-exp-zoomed}, shows stable results for all languages in terms of the curve characteristics.

We further show results in Tables ~\ref{tab:all-error-rates-gamma}, ~\ref{tab:all-error-rates-delta-c4} for C4 dataset, and Table~\ref{tab:all-error-rates-lfqa} for LFQA dataset. In these tables we show the results when fixing the $z$-threshold to 4.0, 5.0 for KGW, Unigram, XSIR, and $p$-value to $10e-5$ and $10e-4$ for EXP method. The same conclusion can be drawn from using larger $\gamma$ values with smaller $\delta$ values. Using larger values of the threshold makes detection harder for those $\gamma$ settings as shown in Table ~\ref{tab:all-error-rates-gamma} for all languages. The effect of detection drop is minimized when LFQA instruction-following task is used as shown in Table~\ref{tab:all-error-rates-lfqa} with exception of Unigram method where TPRs drops are clear for fixed $z$-threshold.

\begin{table*}[th!]
    \centering
    \small
    \caption{Performance Metrics for Multiple Languages and Watermarking Methods calculated by following the test statistics score with two threshold values as shown by $z$. The results show the effect of varying $\gamma$ values with a fixed $\delta=2.0$ for KGW \citet{kirchenbauer2023watermark} and Unigram \citet{zhao2023provable}.}
    \begin{tabular}{ccccccccccc}
        \toprule
        \multirow{2}{*}{Method} & \multirow{2}{*}{Language} & \multirow{2}{*}{$\gamma$} & \multicolumn{4}{c}{Metrics ($z=4.0$)} & \multicolumn{4}{c}{Metrics ($z=5.0$)} \\
        \cmidrule(lr){4-7}
        \cmidrule(lr){8-11}
        & & & TPR & TNR & FPR & FNR & TPR & TNR & FPR & FNR\\
        \midrule
        \multirow{12}{*}{KGW} & \multirow{3}{*}{English}
        & 0.1 & 0.998 & 1.0 & 0.0 & 0.002 & 0.994 & 1.0 & 0.0 & 0.006 \\
        & & 0.5 & 0.998	& 1.0 & 0.0	& 0.002 & 0.998 & 1.0 & 0.0 & 0.002 \\
        & & 0.9 & 0.162	& 1.0 & 0.0 & 0.838 & 0.0 & 1.0 & 0.0 & 1.0 \\
        \cmidrule(lr){2-11}
        & \multirow{3}{*}{Arabic}
        & 0.1 & 0.970 & 0.990 & 0.010 & 0.030 & 0.954 & 0.994 & 0.006 & 0.046 \\
        & & 0.5 & 0.974 & 0.994 & 0.006 & 0.026 & 0.958 & 0.998 & 0.002 & 0.042 \\
        & & 0.9 & 0.112 & 1.0 & 0.0 & 0.888 & 0.0 & 1.0 & 0.0 & 1.0 \\
        \cmidrule(lr){2-11}
        & \multirow{3}{*}{Chinese}
        & 0.1 & 1.0 & 0.992 & 0.008 & 0.0 & 0.994 & 0.992 & 0.008 & 0.006 \\
        & & 0.5 & 1.0 & 0.998 & 0.002 & 0.0 & 0.992 & 1.0 & 0.0 & 0.008 \\
        & & 0.9 & 0.162 & 1.0 & 0.0 &0.838 & 0.0 & 1.0 & 0.0 & 1.0 \\
        \cmidrule(lr){2-11}
        & \multirow{3}{*}{Indonesian}
        & 0.1 & 0.956 & 1.0 & 0.0 & 0.044 & 0.902 & 1.0 & 0.0 & 0.098 \\
        & & 0.5 & 0.966 & 0.996 & 0.004 & 0.034 & 0.862 & 1.0 & 0.0 & 0.138 \\
        & & 0.9 & 0.026 & 1.0 & 0.0 & 0.974 & 0.0 & 1.0 & 0.0 & 1.0 \\
        \midrule
        \multirow{12}{*}{Unigram} & \multirow{3}{*}{English}
        & 0.1 & 0.986 & 0.998 & 0.002 & 0.014 & 0.936 & 1.0 & 0.0 & 0.064 \\
        & & 0.5 & 1.0 & 0.990 & 0.010 & 0.0 & 1.0 & 0.998 & 0.002 & 0.0 \\
        & & 0.9 & 0.244 & 1.0 & 0.0 & 0.756 & 0.0 & 1.0 & 0.0 & 1.0 \\
        \cmidrule(lr){2-11}
        & \multirow{3}{*}{Arabic}
        & 0.1 & 0.976 & 0.970 & 0.030 & 0.024 & 0.948 & 0.976 & 0.024 & 0.052 \\
        & & 0.5 & 0.986 & 0.970 & 0.030 & 0.014 & 0.970 & 0.996 & 0.004 & 0.030 \\
        & & 0.9 & 0.432 & 0.998 & 0.002 & 0.568 & 0.0 & 1.0 & 0.0 & 1.0 \\
        \cmidrule(lr){2-11}
        & \multirow{3}{*}{Chinese}
        & 0.1 & 1.0 & 0.856 & 0.144 & 0.0 & 1.0 & 0.926 & 0.074 & 0.0 \\
        & & 0.5 & 0.998 & 0.876 & 0.124 & 0.002 & 0.996 & 0.926 & 0.074 & 0.004 \\
        & & 0.9 & 0.500 & 0.946 & 0.054 & 0.500 & 0.0 & 1.0 & 0.0 & 1.0 \\
        \cmidrule(lr){2-11}
        & \multirow{3}{*}{Indonesian}
        & 0.1 & 0.952 & 0.970 & 0.030 & 0.048 & 0.922 & 0.982 & 0.018 & 0.078 \\
        & & 0.5 & 0.984 & 0.954 & 0.046 & 0.016 & 0.962 & 0.990 & 0.010 & 0.038 \\
        & & 0.9 & 0.052 & 0.998 & 0.002 & 0.948 & 0.0 & 1.0 & 0.0 & 1.0 \\
        \bottomrule
    \end{tabular}
    \label{tab:all-error-rates-gamma}
\end{table*}

\begin{table*}[th!]
    \centering
    \small
    \caption{Performance Metrics for Multiple Languages and Watermarking Methods calculated by following the test statistics score with fixed $z=4.0$ and automatic $z$ threshold for LFQA dataset. Here we employ $\gamma=0.5$ and $\delta=2.0$ for KGW, Unigram and XSIR.}
    \begin{tabular}{cccccccc}
        \toprule
        \multirow{2}{*}{Method} & \multirow{2}{*}{Language} & \multicolumn{4}{c}{Metrics ($z=4.0$)} & \multicolumn{2}{c}{Automatic $z$ thresholds} \\
        \cmidrule(lr){3-6}
        \cmidrule(lr){7-8}
        & & TPR & TNR & FPR & FNR & TPR@FPR$=0.1\%$ & TPR@FPR$=1\%$ \\
        \midrule
        \multirow{4}{*}{KGW} 
        & English & 0.854 & 1.000 & 0.000 & 0.146 & 0.958 & 0.984\\
        & Arabic & 0.834 & 1.000 & 0.000 & 0.166 & 0.936 & 0.986\\
        & Chinese & 0.942 & 0.998 & 0.002 & 0.058 & 0.918 & 0.998\\
        & Indonesian & 0.972 & 1.000 & 0.000 & 0.028 & 0.990 & 1.000\\
        \midrule
        \multirow{4}{*}{Unigram} 
        & English & 0.490 & 1.000 & 0.000 & 0.510 & 0.626 & 0.944\\
        & Arabic & 0.568 & 1.000 & 0.000 & 0.432 & 0.648 & 0.938\\
        & Chinese & 0.956 & 0.988 & 0.012 & 0.044 & 0.876 & 0.954\\
        & Indonesian & 0.616 & 1.000 & 0.000 & 0.384 & 0.894 & 0.980\\
        \midrule
        \multirow{4}{*}{EXP} 
        & English & 0.980 & 1.000 & 0.000 & 0.020 & 0.990 & 0.992\\
        & Arabic & 0.996 & 1.000 & 0.000 & 0.004 & 1.000 & 1.000\\
        & Chinese & 1.000 & 1.000 & 0.000 & 0.000 & 1.000 & 1.000\\
        & Indonesian & 1.000 & 1.000 & 0.000 & 0.000 & 1.000 & 1.000\\
        \midrule
        \multirow{4}{*}{XSIR} 
        & English & 0.958 & 1.000 & 0.000 & 0.042 & 0.964 & 0.992\\
        & Arabic & 0.994 & 0.920 & 0.080 & 0.006 & 0.866 & 0.976\\
        & Chinese & 0.996 & 0.982 & 0.018 & 0.004 & 0.970 & 0.994\\
        & Indonesian & 1.000 & 0.994 & 0.006 & 0.000 & 0.994 & 1.000\\
        \bottomrule
    \end{tabular}
    \label{tab:all-error-rates-lfqa}
\end{table*}

\begin{table*}[th!]
    \centering
    \small
    \caption{Performance Metrics for Multiple Languages and Watermarking Methods calculated by following the test statistics score for C4 dataset. The results show the effect of varying $\delta$ values with a fixed of $\gamma=0.5$ for KGW, Unigram, and XSIR. Using a threshold of $0.2$ for XSIR resulted in higher FPRs Specifically for the English language.}
    \begin{tabular}{ccccccccccc}
        \toprule
        \multirow{2}{*}{Method} & \multirow{2}{*}{Language} & \multirow{2}{*}{$\delta$} & \multicolumn{4}{c}{Metrics ($z=4.0$)} & \multicolumn{4}{c}{Metrics ($z=5.0$)} \\
        \cmidrule(lr){4-7}
        \cmidrule(lr){8-11}
        & & & TPR & TNR & FPR & FNR & TPR & TNR & FPR & FNR \\
        \midrule
        \multirow{12}{*}{KGW} & \multirow{3}{*}{English}
        & 2.0 & 0.998 & 1.0 & 0.0 & 0.002 & 0.998 & 1.0 & 0.0 & 0.002 \\
        & & 5.0 & 1.0 & 1.0 & 0.0 & 0.0 & 1.0 & 1.0 & 0.0 & 0.0 \\
        & & 10.0 & 1.0 & 1.0 & 0.0 & 1.0 & 1.0 & 1.0 & 0.0 & 0.0 \\
        \cmidrule(lr){2-11}
        & \multirow{3}{*}{Arabic}
        & 2.0 & 0.974 & 0.994 & 0.006 & 0.026 & 0.958 & 0.998 & 0.002 & 0.042\\
        & & 5.0 & 1.0 & 0.998 & 0.002 & 0.0 & 1.0 & 1.0 & 0.0 & 0.0 \\
        & & 10.0 & 1.0 & 0.998 & 0.002 & 0.0 & 1.0 & 1.0  0.0 & 0.0 \\
        \cmidrule(lr){2-11}
        & \multirow{3}{*}{Chinese}
        & 2.0 & 1.0 & 0.998 & 0.002 & 0.0 & 0.992 & 1.0 & 0.0 & 0.008 \\
        & & 5.0 & 1.0 & 0.994 & 0.006 & 0.0 & 1.0 & 0.998 & 0.002 & 0.0 \\
        & & 10.0 & 1.0 & 0.998 & 0.002 & 0.0 & 1.0 & 1.0 & 0.0 & 0.0 \\
        \cmidrule(lr){2-11}
        & \multirow{3}{*}{Indonesian}
        & 2.0 & 0.966 & 0.996 & 0.004 & 0.034 & 0.862 & 1.0 & 0.0 & 0.138 \\
        & & 5.0 & 0.996 & 1.0 & 0.0 & 0.004 & 0.992 & 1.0 & 0.0 & 0.008 \\
        & & 10.0 & 1.0 & 1.0 & 0.0 & 0.0 & 1.0 & 1.0 & 0.0 & 0.0 \\
        \midrule
        \multirow{12}{*}{Unigam} & \multirow{3}{*}{English}
        & 2.0 & 1.0 & 0.990 & 0.010 & 0.0 & 1.0 & 0.998 & 0.002 & 0.0 \\
        & & 5.0 & 1.0 & 0.984 & 0.016 & 0.0 & 1.0 & 1.0 & 0.0 & 0.0 \\
        & & 10.0 & 1.0 & 0.984 & 0.016 & 0.0 & 1.0 & 1.0 & 0.0 & 0.0 \\
        \cmidrule(lr){2-11}
        & \multirow{3}{*}{Arabic}
        & 2.0 & 0.986 & 0.970 & 0.030 & 0.014 & 0.97 & 0.996 & 0.004 & 0.03 \\
        & & 5.0 & 1.0 & 0.972 & 0.028 & 0.0 & 1.00 & 0.990 & 0.010 & 0.00 \\
        & & 10.0 & 1.0 & 0.960 & 0.040 & 0.0 & 1.00 & 0.990 & 0.010 & 0.00 \\
        \cmidrule(lr){2-11}
        & \multirow{3}{*}{Chinese}
        & 2.0 & 0.998 & 0.876 & 0.124 & 0.002 & 0.996 & 0.926 & 0.074 & 0.004 \\
        & & 5.0 & 1.0 & 0.890 & 0.110 & 0.0 & 1.0 & 0.924 & 0.076 & 0.0 \\
        & & 10.0 & 1.0 & 0.878 & 0.122 & 0.0 & 1.0 & 0.922 & 0.078 & 0.0 \\
        \cmidrule(lr){2-11}
        & \multirow{3}{*}{Indonesian}
        & 2.0 & 0.984 & 0.954 & 0.046 & 0.016 & 0.962 & 0.990 & 0.010 & 0.038 \\
        & & 5.0 & 1.0 & 0.962 & 0.038 & 0.0 & 1.0 & 0.994 & 0.006 & 0.0 \\
        & & 10.0 & 1.0 & 0.942 & 0.058 & 0.0 & 1.0 & 0.984 & 0.016 & 0.0 \\
        \midrule
        & & & \multicolumn{4}{c}{Metric ($z=0.2$)} & \multicolumn{4}{c}{Metric ($z=0.3$)} \\
        \cmidrule(lr){4-7}
        \cmidrule(lr){8-11}
        & & & TPR & TNR & FPR & FNR & TPR & TNR & FPR & FNR \\
        \cmidrule(lr){2-11}
        \multirow{12}{*}{XSIR} & \multirow{3}{*}{English}
        & 2.0 & 1.0 & 0.702 & \textbf{0.298} & 0.0 & 0.994 & 0.958 & 0.042 & 0.006 \\
        & & 5.0 & 1.0 & 0.734 & \textbf{0.266} & 0.0 & 0.998 & 0.964 & 0.036 & 0.002 \\
        & & 10.0 & 0.996 & 0.708 & \textbf{0.292} & 0.004 & 0.996 & 0.960 & 0.040 & 0.004 \\
        \cmidrule(lr){2-11}
        & \multirow{3}{*}{Arabic}
        & 2.0 & 0.998 & 0.916 & 0.084 & 0.002 & 0.996 & 0.994 & 0.006 & 0.004 \\
        & & 5.0 & 1.0 & 0.928 & 0.072 & 0.0 & 0.998 & 0.994 & 0.006 & 0.002 \\
        & & 10.0 & 1.0 & 0.898 & \textbf{0.102} & 0.0 & 0.998 & 0.982 & 0.018 & 0.002 \\
        \cmidrule(lr){2-11}
        & \multirow{3}{*}{Chinese}
        & 2.0 & 1.0 & 0.966 & 0.034 & 0.0 & 0.994 & 0.994 & 0.006 & 0.006 \\
        & & 5.0 & 0.996 & 0.944 & 0.056 & 0.004 & 0.992 & 0.982 & 0.018 & 0.008 \\
        & & 10.0 & 0.992 & 0.964 & 0.036 & 0.008 & 0.992 & 0.994 & 0.006 & 0.008 \\
        \cmidrule(lr){2-11}
        & \multirow{3}{*}{Indonesian}
        & 2.0 & 0.988 & 0.990 & 0.010 & 0.012 & 0.976 & 0.998 & 0.002 & 0.024 \\
        & & 5.0 & 0.998 & 0.994 & 0.006 & 0.002 & 0.998 & 1.0 & 0.0 & 0.002 \\
        & & 10.0 & 0.996 & 0.984 & 0.016 & 0.004 & 0.994 & 0.998 & 0.002 & 0.006 \\
        \midrule
        & & & \multicolumn{4}{c}{Metric ($p=10e-5$)} & \multicolumn{4}{c}{Metric ($p=10e-4$)} \\
        \cmidrule(lr){4-7}
        \cmidrule(lr){8-11}
        & & & TPR & TNR & FPR & FNR & TPR & TNR & FPR & FNR \\
        \cmidrule(lr){2-11}
        \multirow{4}{*}{EXP} & English
        & - & 0.998 & 1.0 & 0.0 & 0.002 & 0.998 & 0.996 & 0.004 & 0.002 \\
        & Arabic & - & 0.982 & 1.0 & 0.0 & 0.018 & 0.982 & 1.0 & 0.0 & 0.018 \\
        & Chinese & - & 0.994 & 1.0 & 0.0 & 0.006 & 0.994 & 1.0 & 0.0 & 0.006 \\
        & Indonesian & - & 0.990 & 1.0 & 0.0 & 0.010 & 0.992 & 1.0 & 0.0 & 0.008\\
        \bottomrule
    \end{tabular}
    \label{tab:all-error-rates-delta-c4}
\end{table*}

\subsection{Watermark Detection After Attacks for Syntactical Methods}
In Figure~\ref{fig:roc-curve-kgw-unigram-exp-all-attacks} shows the completion of attack pipeline performed on KGW, Unigram, and EXP. Translation followed by paraphrase attacks are no worse than translations alone for syntactical methods. However, we notice that beginning with English in the pipeline performs better in all methods as can be seen from the top row of all the figures. Translation-paraphrase-translation performs well when English is the source language in all methods across all target languages, whereas the detection is very low when English is not the source language.

\section{GPT-Judger Fairness Experiments}\label{sec:appendix-gpt-fairness}
We perform two experiments in an attempt to elicit any possible biases toward a specific language over the other. First, we focus on the languages we study which are English, Arabic, Chinese and Indonesian. We also add closely related languages such as Persian or Farsi (close to Arabic), German (close to English), and Japanese (close to Chinese). To prepare our data with ground truth values, we use natural English examples from the C4 dataset then we translate these examples to all other languages using a powerful GPT model such as GPT-3.5-Turbo. The objective here is to ensure that we have the same text but in different languages to ascertain any biases toward a specific language. In all of our judging experiments we randomized the order in which the texts are fed to the LLM and use two runs under different seeds to allow for a more rigor investigation.

\noindent\textbf{Judging Texts from Different Languages.} To further analyze the judger's result, we conduct a fairness experiment we call "Translation Experiment" in which we investigate whether GPT-Judger shows bias toward a specific language when given the same text in multiple languages. We use the same judge we used for our main experiment in the main content of the paper, which means we used the same LLM and system prompt with slight adjustment to the prompt to account for multiple options instead of only two options (unwatermarked and watermarked texts.) Table~\ref{tab:trans-verdict-percentages} shows the result of judging the same text in multiple languages. The \emph{hard win} column reflects the judge final verdict for the language of the winning text. the \emph{first-last} column reflects the percentage of the winning text when it is placed in the first or the last option in the prompt to investigate position bias if any ~\citep{pezeshkpour2023large}. As can be seen from the table, we see $6.5\%$ for both English and Chinese, yet their hard win rates are largely different. Therefore, it is not clear whether the order affects the final verdict of the judge in this settings. However, it's clear that some languages are preferred over other despite ensuring that the texts are mere \emph{translations} of each other, and therefore should have more TIE percentages \footnote{We instruct the judge to clearly return TIE as the final verdict if the text quality of all languages are equally good.}. For example, German language is the most preferred one, followed by Arabic, and finally English, while Chinese, Persian, and Indonesian languages receives the lowest scores. Persian language, despite the fact that it originates from the same family as Arabic, receives the lowest scores. Therefore, we hypothesize that there is a bias toward languages that the judge is most proficient at or more familiar with.

\begin{table*}[th]
    \centering
    \small
    \setlength{\tabcolsep}{4pt}
    \renewcommand\arraystretch{0.9}
    \caption{Translation Verdict Percentages over two runs with different seeds. Each run contains 100 samples from 500 translation examples for a total of 200 examples. The \emph{first-last} column reflects the percentage of the winning text when it is placed in the first or the last option in the prompt to investigate position bias if any.}
    \begin{tabular}{llc}
        \toprule
        \textbf{Language} & \textbf{hard win (\%)} & \textbf{first-last (\%)} \\
        \midrule
        EN & 19.50 $\pm$ 4.95 & 6.50 \\
        JA & 14.50 $\pm$ 3.54 & 13.00 \\
        FA & 2.00  $\pm$ 0.00 & 2.00  \\
        AR & 23.00 $\pm$ 5.66 & 13.00 \\
        ZH & 7.00  $\pm$ 1.41 & 6.50  \\
        DE & 30.50 $\pm$ 3.54 & 11.00 \\
        ID & 3.50  $\pm$ 0.71 & 3.50  \\
        \midrule
        Total & 100 & 55.5 \\
        \bottomrule
    \end{tabular}
    \label{tab:trans-verdict-percentages}
\end{table*}

\begin{table*}[th]
    \centering
    \caption{Paraphrase Verdict Percentages over two runs with different seeds. Each run contains 100 samples from 500 paraphrased examples from the original non-perturbed text. The Perturbed Text column is the paraphrased version of the Natural Text.}
    \begin{tabular}{lcccc}
        \toprule
        \textbf{Language} & \textbf{Perturbed Text (\%)} & \textbf{Natural Text (\%)} & \textbf{TIE (\%)} & \textbf{Model Failure (\%)} \\
        \midrule
        ZH & 30.00 $\pm$ 1.41 & 26.00 $\pm$ 0.00 & 44.00 $\pm$ 1.41 & 0.00 \\
        FA & 39.50 $\pm$ 3.54 & 20.00 $\pm$ 4.24 & 40.50 $\pm$ 0.71 & 0.00 \\
        JA & 34.00 $\pm$ 7.07 & 28.50 $\pm$ 0.71 & 37.50 $\pm$ 6.36 & 0.00 \\
        AR & 40.50 $\pm$ 4.95 & 30.50 $\pm$ 6.36 & 29.00 $\pm$ 1.41 & 0.00 \\
        DE & 47.50 $\pm$ 6.36 & 28.50 $\pm$ 2.12 & 23.00 $\pm$ 8.49 & 1.00 \\
        ID & 48.00 $\pm$ 4.24 & 30.50 $\pm$ 7.78 & 21.50 $\pm$ 3.54 & 0.00 \\
        EN & 75.50 $\pm$ 9.19 & 24.00 $\pm$ 9.90 & 0.50 $\pm$ 0.71 & 0.00 \\
        \bottomrule
    \end{tabular}
    \label{tab:para-verdict-percentages}
\end{table*}

\noindent\textbf{{Judging Texts from the Same Language}}
In this experiment which we call "Paraphrase Experiment", we used the same judge as the one used watermarking experiments. Our natural text in this case represent the original translated text (for non-English) from the translation experiment. In this experiment, the two texts (natural and perturbed) are essentially the same. In other words, we use GPT-3.5-turbo to paraphrase the original text and we call the resultant text the perturbed text. We create 500 examples in this manner and we used two seeds to sample different 100 examples for our fairness assessments. Table ~\ref{tab:para-verdict-percentages} shows the results from using GPT-Judger with these modified texts to assess their quality. Comparing the paraphrase results with that of the translation, we see a pattern that confirms the translation experiments. We noticed that the languages that received lower scores in translation experiment are at the top of the paraphrase tables, and those that received higher scores are at the bottom in terms of the TIE scores. However, there is an exception for the Indonesian language, which appears to have lower TIE scores. We could argue here that with the model being able to say two texts are equal in quality is not because it is able to assert that the quality of the two texts are equally good, but rather because it might not be proficient in these languages (as seen in Table~\ref{tab:trans-verdict-percentages}) that it preferred to choose a TIE. For Indonesian language, token bias as investigated by~\citet{zheng2309large} can play a role since the alphabets used in this language are the same as the English alphabets. Another important observation is that the perturbed text is favored more by the LLM judger in all the languages as shown in the second column of the table, whereas natural text is less favored by the judger despite not being gone through a lot of modification. This confirms studies that indicate LLMs favoring texts generated by their own model variants ~\citep{ye2024justice} since our perturbed text has gone through perturbations of the same model variant. Therefore, it has become easier for the GPT-judge to select as the winning text. However, in our main watermarking experiments, our perturbed texts (watermarked texts and non-English language texts) don't have such perturbation footprint, which can put aside the bias toward model-own generations from our main experiments.

\end{document}